\documentclass[lettersize,journal]{IEEEtran}
\usepackage{amsmath,amsfonts}
\usepackage{algorithmic}
\usepackage{array}
\usepackage[caption=false,font=normalsize,labelfont=sf,textfont=sf]{subfig}
\usepackage{textcomp}
\usepackage{stfloats}
\usepackage{caption}
\usepackage{subcaption}
\usepackage{url}
\usepackage{verbatim}
\usepackage{graphicx}
\usepackage{cite}
\usepackage{amssymb}
\usepackage{amsthm}
\newtheorem{assumption}{Assumption}
\usepackage{upgreek}
\usepackage{bm}
\usepackage[ruled,vlined,linesnumbered]{algorithm2e}
\usepackage{graphicx}
\usepackage{dsfont}
\usepackage{enumitem}
\usepackage{multirow,booktabs}
\usepackage{tabularray}
\usepackage{color}
\setlist[itemize]{leftmargin=*}
\usepackage{tabularx}
\usepackage{makecell}
\usepackage{booktabs}
\newcolumntype{L}{>{\raggedright\arraybackslash}X}

\let\oldnl\nl% Store \nl in \oldnl
\newcommand{\nonl}{\renewcommand{\nl}{\let\nl\oldnl}}% Remove line number for one line

\SetArgSty{textnormal}  

\SetKw{KwRet}{return}
\SetKw{KwAll}{all}
\SetKw{KwNot}{not}

\SetKwInOut{Input}{Input}
\SetKwInOut{Output}{Output}
\SetKwInOut{Data}{Data}
\SetKwInOut{Result}{Result}
\SetKwInOut{Stop}{Stop}
\SetKwFunction{fLHS}{latinHypercubeSample}
\SetKwProg{fn}{Function}{:}{\KwRet}

\newtheorem{proposition}{Proposition}

\theoremstyle{definition}
\newtheorem{definition}{Definition}%[section]

\theoremstyle{remark}

\begin{document}

\title{SparseST: Exploiting Data Sparsity in Spatiotemporal Modeling and Prediction}

\author{Junfeng Wu, Hadjer Benmeziane, Kaoutar El Maghraoui, Liu Liu, Yinan Wang
        % <-this % stops a space
        % <-this % stops a space
\IEEEcompsocitemizethanks{
\IEEEcompsocthanksitem Junfeng Wu and Yinan Wang are with the Department of Industrial and Systems Engineering, Rensselaer Polytechnic Institute, Troy, NY, 12180. \protect\\
E-mail: \{wuj26, wangy88\}@rpi.edu
\IEEEcompsocthanksitem Liu Liu is with the Department of Electrical, Computer, and Systems Engineering and the Department of Computer Science, Rensselaer Polytechnic Institute, Troy, NY, 12180. \protect\\
E-mail: liu.liu@rpi.edu
\IEEEcompsocthanksitem Hadjer Benmeziane and Kaoutar El Maghraoui are with the IBM Research.\protect\\
E-mail: Hadjer.Benmeziane@ibm.com, kelmaghr@us.ibm.com}% <-this % stops an unwanted space

\thanks{Manuscript received XXX; revised YYY. (Corresponding Author: Yinan Wang)}

\markboth{Manuscript}}

\maketitle

\begin{abstract}
Spatiotemporal data mining (STDM) has a wide range of applications in various complex physical systems (CPS), i.e., transportation, manufacturing, healthcare, etc. Among all the proposed methods, the Convolutional Long Short-Term Memory (ConvLSTM) has proved to be generalizable and extendable in different applications and has multiple variants achieving state-of-the-art performance in various STDM applications. However, ConvLSTM and its variants are computationally expensive, which makes them inapplicable in edge devices with limited computational resources. With the emerging need for edge computing in CPS, efficient AI is essential to reduce the computational cost while preserving the model performance. Common methods of efficient AI are developed to reduce redundancy in model capacity (i.e., model pruning, compression, etc.). However, spatiotemporal data mining naturally requires extensive model capacity, as the embedded dependencies in spatiotemporal data are complex and hard to capture, which limits the model redundancy. Instead, there is a fairly high level of data and feature redundancy that introduces an unnecessary computational burden, which has been largely overlooked in existing research. Using the sequence of images as an example, (1) the informative pixels at each frame are usually spatially sparse (i.e., only foreground pixels contain meaningful information in action recognition) and (2) the informative features in the first-order difference of two adjacent images are possibly sparse with a fixed or slowly evolving background. Therefore, we developed a novel framework SparseST, that pioneered in exploiting data sparsity to develop an efficient spatiotemporal model. In addition, we explore and approximate the Pareto front between model performance and computational efficiency by designing a multi-objective composite loss function, which provides a practical guide for practitioners to adjust the model according to computational resource constraints and the performance requirements of downstream tasks.

\end{abstract}

\def\abstractname{Note to Practitioners}
\begin{abstract}
This paper is motivated by the need to reduce the computational cost on edge devices with limited computational resources. In various complex physical systems (CPS), spatiotemporal data mining is an important task and naturally requires extensive model capacity, as the embedded dependencies in spatiotemporal data are complex and hard to capture. Instead of model compression, we aim to trade the redundant information in the input data or features for computational efficiency. In addition, our framework provides practitioners with direct control over their preference for efficiency and model accuracy. And the approximated Pareto front of the trade-off between efficiency and accuracy gives the guideline to adjust the model according to computational resource constraints and the performance requirements of downstream tasks.
\end{abstract}


\section{Introduction}

Spatiotemporal data mining (STDM) focuses on data with both spatial and temporal dimensions, capturing how features evolve over time and space, with a wide prevalence in real-world applications such as video analytics, traffic flow forecasting \cite{zhang2024adaptive}, and industrial process monitoring \cite{lee2014recent}. However, deep neural networks (DNN) for the application of STDM are computationally expensive in advanced manufacturing settings, including but not limited to process monitoring, quality control, and anomaly detection. With the increasing amount of data provided by different sensors in smart manufacturing systems, the size and complexity of DNN models are also increased, which means deploying DNN models is more and more computationally expensive. However, most edge computing or embedded devices in manufacturing systems only have limited memory, storage, and computational capacity, leading to the pressing need to improve the computational efficiency of DNN models while preserving their performance. To handle the issue mentioned above, efficient AI methods are developed to reduce the computational cost and redundancy in DNN models \cite{cheng2017survey, choudhary2020comprehensive}.

% From the perspective of reducing model redundancy, the common method of efficient AI is model compression \cite{cheng2017survey, choudhary2020comprehensive}, including weight pruning, quantization, low-rank factorization, etc. These methods reduce computational costs and memory requirements by pruning model weights, storing model weights with a lower precision, and approximating model weights with low-rank components, respectively. Another perspective of efficient AI is to leverage the sparsity in the raw input data or intermediate features \cite{neil2017delta, graham2017submanifold, ren2018sbnet, pan2018compressing} to reduce the computational cost. For example, in a matrix-vector multiplication (MxV) between a weight matrix and a sparse vector, multiplications involving zero elements can be skipped since they do not contribute to the output. 

We aim to develop efficient AI methods to accelerate the classical Convolutional Long Short-Term Memory (ConvLSTM) model for the downstream task of spatiotemporal prediction and anomaly detection. ConvLSTM was proposed by Shi et al. \cite{shi2015convolutional}, which integrates convolutional operations into the LSTM architecture. Unlike traditional LSTM models that use fully connected layers in gate computation, ConvLSTM employs convolution to capture the spatial patterns of input data and hidden state features, effectively combining the strengths of both Convolutional Neural Network (CNN) and LSTM to better handle spatiotemporal data. 

ConvLSTM models are computationally expensive, as they integrate convolutional operations into recurrent architectures. Applying existing model compression techniques to reduce model redundancy of ConvLSTM is nontrivial for several reasons: (1) Recurrent Neural Network (RNN) based model is susceptible to vanishing and exploding gradient \cite{jain2022survey}, and aggressive pruning or quantization can further deteriorate this issue, making it more difficult to train the model; (2) low-rank factorization methods introduce high computational overhead for decomposition and impose strict structural constraints on the factorized components, which can limit the representational capacity of the compressed model \cite{yin2020compressing}.
Rather than directly compressing the model, which can compromise training stability and limit representation capacity, we adopt a complementary perspective: reducing redundancy by exploiting the inherent sparsity of input data and intermediate features. This strategy improves efficiency while preserving the integrity of the ConvLSTM architecture.

We consider data with high spatial sparsity, such as images that contain only a small portion of informative pixels within a mostly black background. Such spatially sparse data can be efficiently processed by the 2D Sparse Convolution \cite{graham2017submanifold}, which discards Multiply–Accumulate (MAC) operations involving zero elements.  However, this method has two main drawbacks: (1) 2D Sparse Convolution can not exploit temporal sparsity in spatiotemporal data. For example, the first-order difference of pixel values between two adjacent image frames is often sparse due to slow temporal change; (2) 2D Sparse Convolution can cause a feature dilation effect, which means feature sparsity decreases as model depth increases \cite{graham2017submanifold}, thereby reducing computational efficiency in deeper layers. To address the above limitations, we adopt the Delta Network (DN) Algorithm \cite{neil2017delta} to leverage temporal sparsity by applying a threshold to zero out feature values with a small first-order temporal difference. Therefore, it effectively leverages the temporal redundancy in spatiotemporal data and mitigate the problem of feature dilation in deeper layers. 

In addition to the importance of efficiency in spatiotemporal modeling, there is a pressing need for practitioners to understand how to control the trade-off between efficiency and model performance, enabling them to adjust this balance according to computational resource constraints and performance requirements of downstream tasks. To address this, we formulate our problem as a multi-objective Pareto front learning task. Specifically, we design a multi-objective composite loss for two conflicting objectives: performance and efficiency, and train the neural network using the Smooth Tchebycheff Scalarization (STCH) method \cite{lin2024smooth}, which can effectively explore Pareto optimal points in both convex and non-convex regions of the Pareto front. Finally, we train a multi-task Gaussian Process (GP) to jointly model the dependency between model performance and efficiency. By approximating the Pareto front, we provide a guide for practitioners to fine-tune the preference weight and achieve a desired trade-off.

In this work, we propose SparseST, a framework designed as an efficient variant of ConvLSTM, which leverages the spatial and temporal sparsity present in raw input data and intermediate feature representations. In the meantime, we investigate the trade-off between model performance and efficiency by approximating the Pareto front. Our experimental results demonstrate notable computational savings with respect to (w.r.t.) Floating Point Operations (FLOPs) compared with baseline models on spatiotemporal prediction and anomaly detection tasks. The approximated Pareto front further shows the trade-off trend as expected. We summarize the contribution of this work as follows:
\begin{enumerate}
    \item We propose SparseST, a framework that integrates 2D Sparse Convolution and the DN algorithm into the ConvLSTM architecture. By exploiting both data and feature sparsity, SparseST accelerates spatiotemporal modeling while preserving model performance.
    
    \item We formulate the trade-off between model performance and efficiency as a multi-objective optimization problem. By approximating the Pareto front, our framework provides practical guidelines for practitioners to adjust preference weights based on computational resource constraints and downstream performance requirements.  

\end{enumerate}

The remainder of this paper is organized as follows. Section \ref{sec: 2} reviews related literature on efficient AI methods and multi-objective Pareto front learning. Section \ref{sec: 3} presents the preliminary background on the ConvLSTM model and the spatially sparse tensor. Section \ref{sec: 4} demonstrates the proposed SparseST from the concepts of Spatially Sparse Convolution and the DN Algorithm to the full model architecture, computational cost analysis, and the design of the multi-objective composite loss. Section \ref{sec: 5} shows the experimental details, including dataset descriptions, hyperparameter settings, evaluation metrics, and results analysis for the two downstream tasks. Finally, Section \ref{sec: 6} concludes this work and discusses the limitations.

\section{Literature Review} \label{sec: 2}

This section reviews the literature on efficient AI from the perspectives of model compression and data or feature redundancy. Particularly, the related work on Multi-objective Pareto Front Learning is summarized, which is crucial to developing the SparseST framework. We also compare our method with the existing work to show the new methodological contribution.

\subsection{Efficient AI}

Training a deep ConvLSTM model requires a massive number of MAC operations in each layer, both in forward and backward propagation. Model compression is a widely used method to reduce MAC operations for better efficiency, including pruning, quantization, low-rank factorization, etc. Another approach is to reduce MAC operations by exploiting the intrinsic spatial and temporal sparsity of sequential data.

\subsubsection{Model Compression}

\paragraph{Pruning}

Pruning eliminates those redundant or less important model parameters that do not contribute much to model performance, therefore reducing storage size, computational cost, and energy consumption without a substantial loss in model accuracy.

There are various pruning methods, such as weight pruning, neuron pruning, layer pruning, etc. Han et al. \cite{han2015learning} proposed an iterative process of pruning and retraining. In each iteration, the model is first trained to learn informative weights. Then, unimportant weights with small magnitudes are pruned. Finally, the pruned model is retrained to fine-tune the pruned weights and regain model performance. However, the majority of parameters removed by this method are from the fully connected layers, where the computational cost is much lower than that of the convolutional layers \cite{li2017pruning}. To effectively compress CNN layers, Li et al. follow the same iterative pruning and retraining process as \cite{han2015learning} and proposed a new method to prune unimportant convolutional filters evaluated by the summation of their absolute weights \cite{li2017pruning}. Similarly, Liu et al. \cite{liu2017learning} proposed a network slimming method for pruning the entire channel of CNN by applying a sparsity-inducing $l_1$ regularizer to the scaling factors in batch normalization (\textbf{BN}) layers, effectively identifying and pruning less important channels. 

Pruning is also explored in RNN models. Narang et al. \cite{narang2017exploring} pruned the weights of RNNs during the model training and achieved a weight sparsity of $90\%$ with a small loss in accuracy by multiplying each weight matrix with a binary mask at each update step. At regular intervals, the masks are updated by a threshold function controlled by a set of hyperparameters. This technique does not require additional retraining steps used in \cite{han2015learning}. However, it introduced an additional step of hyper-parameter tuning to control the threshold, which is difficult to implement in practice \cite{narang2017exploring}. Wen et al. \cite{wen2018learning} explored the structured sparsity of LSTMs and proposed the concept of Intrinsic Sparse Structures (ISS). They novelly identified the basic structure inside RNNs, which can be considered as a group, thus making it possible to remove weights independently across all the basic structures within ISS.

\paragraph{Quantization}

Quantization is a model compression technique to reduce the number of bits required for the representation and storage of model weights. For example, weights in 32-bit floating-point full precision representation can be quantized to a lower precision representation, such as 16-bit or 8-bit. Ott \cite{ott2016recurrent} designed multiple experiments to test different methods to quantize the weights of three major RNN types: vanilla RNN, GRU, and LSTM, showing for the first time how low-precision quantization of weights can be performed during RNN training. Besides weight quantization, gradient, and neuron activation can also be quantized to reduce the network size and memory requirement during model training and inference. Wei et al. \cite{wen2017terngrad} introduce TernGrad, a method for gradient quantization that reduces the communication cost in distributed deep learning by quantizing gradients to ternary values ${(-1, 0, 1)}$. This approach significantly reduces the amount of data that needs to be communicated between workers and the parameter server during training while maintaining high accuracy. Hubara et al. \cite{hubara2018quantized} quantized both weights and neuron activations of vanilla LSTM and CNN models, achieving low-precision training and inference without significant loss in accuracy.

\paragraph{Low-rank factorization}

The low-rank factorization method aims to use matrix or tensor decomposition to estimate the informative parameters of the DNNs. For example, A matrix $W$ of size $m \times n$ can be approximated by the product of two smaller matrices $A$ and $B$, where $A$ is of size $m \times k$ and $B$ is of size $k \times n$, with $k<\min (m, n)$. This reduces the number of parameters from $m n$ to $m k+k n$. Common factorization techniques for 2D matrices include Singular Value Decomposition (SVD), Principal Component Analysis (PCA), and their variant \cite{cheng2017survey}\cite{denil2013predicting}\cite{lu2017fully}. For high-dimensional tensors in the case of convolutional filters, factorization methods such as Canonical Polyadic (CP) decomposition, Batch Normalization (BN) low-rank decomposition, etc. \cite{lebedev2014speeding}\cite{tai2015convolutional}. Wang et al. \cite{wang2022tensor} proposed a CPAC-Conv layer by adopting CP-decomposition to compress convolutional kernels, with formulas for both forward and backward propagations derived. The value of the decomposed kernels indicates significant feature maps, which are informative for feature selection. Lu et al. \cite{lu2016learning} first undertook a systematic study and investigated redundancies in recurrent architectures such as RNN and LSTM. They compared the compression performance for different low-rank factorization methods \cite{sak2014long} \cite{kingsbury2016low} and found that a hybrid strategy of using structured matrices in the bottom layers and shared low-rank factors on the top layers is particularly effective \cite{lu2016learning}. Later advancement shows various tensor decomposition-based compression methods, including tensor train \cite{yang2017tensortrain}, tensor ring \cite{pan2018compressing}, and block-term \cite{ye2018learning}, can bring several orders of magnitude fewer parameters for large-scale RNNs while still maintaining high classification/prediction performance. However, methods mentioned above suffer from high computational overhead and limited representation ability of compressed RNN models due to extra flatten or permutation operations and strict constraints on either the shapes or the combination manners of the low-rank tensor components \cite{yin2020compressing}.

Although the model compression methods mentioned above have achieved prominent compression ratios, they only reduced the computational cost from the perspective of model redundancy, but didn't exploit the data or feature redundancy. Besides, applying existing model compression techniques to ConvLSTM is problematic due to issues such as unstable gradients, dimension mismatch, extensive retraining, degraded model capacity, etc. The following methods, which leverage the spatiotemporal sparsity of input data and intermediate features, will not suffer from the above issues, yet are still efficient in reducing the computational costs without sacrificing the model performance.

\subsubsection{Data and Feature Sparsity}

\paragraph{Spatially Sparse Convolution}

Spatially Sparse Convolution is a convolution operation optimized for processing spatially sparse data, for example, 3D point cloud data obtained from a LiDAR scanner, images of handwritten numbers in a sparse background with mostly zero pixel values, etc. Note that Spatially Sparse Convolution is different from the conventional dense CNN with weight pruning or compression. The word ``sparse'' only means sparsely distributed informative entries in data tensors. 

Graham et al. \cite{graham2014spatiallysparse} first introduced 2D Sparse Convolution for handwriting recognition and image classification tasks. A handwritten character can be represented as a sparse matrix, in which only pixels indicating the character are informative. Instead of conducting convolution operations over the entire sparse matrix (i.e., conventional dense CNN), 2D Sparse Convolution only conducts convolution operations over nonzero features (i.e., pixels indicating characters) and skips those pixels that only represent backgrounds (i.e., usually are zeros). Therefore, 2D Sparse Convolution significantly reduces computational cost by exploiting the sparsity in matrix-like data. Afterward, a sparse 3D CNN is explored to efficiently process 3D point cloud data \cite{engelcke2017vote3deep}\cite{graham2015sparse}. However, the current Sparse Convolution still dilates the number of nonzero values over layers of convolution operations, which reduces the level of sparsity of feature matrices over the layers of the neural network.

In \cite{graham20183d} \cite{graham2017submanifold}, Graham et al. introduced Submanifold Sparse Convolution on a 3D point cloud data segmentation task. This method only conducts the convolution with the corresponding receptive field centered at a nonzero element, thereby keeping the same sparsity pattern throughout the layers of the network without dilating the feature maps. Later, Yan \cite{yan2018second} adopted Submanifold Sparse Convolution \cite{graham2017submanifold} for the 3D object detection task and optimized the data flow to make it more GPU-friendly compared with \cite{graham2017submanifold}. In the more recent work by \cite{tang2022torchsparse} and \cite{tang2023torchsparse++}, Tang et al. propose the TorchSparse library for processing 3D point cloud data in the application of autonomous driving, with optimized computational regularity and data movement compared to previous work.

In summary, although Spatially Sparse Convolution is well studied, especially in 3D object detection and segmentation tasks, to the best of our knowledge, there are still significant gaps to directly apply it to STDM, which includes (1) Submanifold Sparse Convolution is inherently inapplicable to STDM since it keeps the same sparsity pattern across layers, which hinders the model to learn about temporal dynamics of spatiotemporal data; (2) The dilation issue of Sparse Convolution results in degraded computational efficiency in deeper layers; (3) Sparse Convolution can only efficiently process spatially sparse data (with a sparse ground), not applicable to spatiotemporal data with a fixed but not sparse background. To mitigate the issues mentioned above, we resort to the DN Algorithm. 

\paragraph{Delta Network Algorithm}

The DN Algorithm, first proposed by Neil et al. \cite{neil2017delta}, is a computational reuse method to induce temporal sparsity of sequential data and intermediate features in the RNN model. It sets a delta threshold to zero out elements below that threshold, resulting in sparse delta vectors. Replacing the dense state vectors with sparse delta vectors, which contain the temporal difference of the states between two adjacent time steps, can reduce both computational costs and required memories \cite{gao2022spartus}. Gao et al. \cite{gao2018deltarnn} first adopted the DN Algorithm to accelerate the Gated Recurrent Unit (GRU) on a specifically designed FPGA hardware. The following work by Gao et al. \cite{gao2020edgedrnn} further reduced the memory requirement of edge computing devices by storing RNN parameters in off-chip memory and explored the trade-off of model performance and sparsity level under different delta threshold settings. However, this work only exploited temporal data sparsity. Later, Gao et al. \cite{gao2022spartus} proposed Spartus, a hardware accelerator for LSTM models using a structured pruning method with the DN algorithm, making it possible to implement the LSTM model on the edge device with small memory for real-time online speech recognition tasks.

The existing work of applying the DN Algorithm to RNN models is about the co-design of hardware and algorithms with applications on speech recognition and audio digit recognition tasks. However, it has never been used to explore the temporal correlation of spatiotemporal data. More importantly, the DN algorithm is a perfect match for Sparse Convolution to mitigate the dilation issue and relax the constraint of spatially sparse data (with a sparse background)  to a fixed and dense background.

\subsection{Multi-objective Pareto Front Learning} \label{sec: 2.2}

Learning the Pareto front is a fundamental task in multi-objective optimization, which aims to recover the full set of all Pareto optimal solutions illustrating the trade-off between conflicting objectives. Optimizing the linear scalarization of multiple objectives (also referred to as the weighted sum method) is a widely used technique to find Pareto optimal solutions. The expression of linear scalarization is shown as follows.

\begin{equation}
    \min _{\boldsymbol{x} \in \mathcal{X}} \sum_{i=1}^m w_i f_i(\boldsymbol{x}),
\label{eq:weighted sum}
\end{equation}

where $w_i$ is the preference weight for the $i^{\text{th}}$ objective $f_i$. 

It converts the multi-objective optimization problem into a single-objective one by weighting the objectives using the preference vector. However, linear scalarization fails to reach Pareto optimal points on the non-convex part of the Pareto front \cite{boyd2004convex}. Therefore, Tchebycheff Scalarization (TCH) \cite{bowman1976relationship, steuer1983interactive} is developed for scalarization to find Pareto optimal points for both convex and non-convex parts of the Pareto front, i.e. 
\begin{equation}
    \min _{\boldsymbol{x} \in \mathcal{X}} \max _{1 \leq i \leq m}\left\{w_i\left(f_i(\boldsymbol{x})-z_i^*\right)\right\}
\label{eq:original TCH}
\end{equation}

where $m$ is the number of objectives, $z_i^*$ is the optimal value for the $i$ th objective.
There is also a promising necessary and sufficient condition for TCH scalarization to find all Pareto optimal solutions \cite{choo1983proper}. Despite its long-standing recognition in the multi-objective optimization community, TCH is rarely used for gradient-based optimization due to the nonsmoothness of the $max$ function in Equation \ref{eq:original TCH}. Consequently, even when all objective functions $f_i$ are differentiable, the TCH formulation remains non-differentiable, making it infeasible for gradient-based optimization \cite{lin2024smooth}. For example, Parallel Efficient Global Optimization (ParEGO) \cite{knowles2006parego} is one of the multi-objective Bayesian optimization (MOBO) methods that employs Tchebycheff scalarization to approximate the Pareto front, which cannot be directly optimized by gradient-based methods. Therefore, Lin et al. \cite{lin2024few, lin2024smooth} proposed a smoothed variant of Tchebycheff scalarization (STCH) for gradient-based multi-objective optimization, making it well-suited for integration with the training of modern neural networks using gradient-based optimizers such as Adam \cite{kingma2014adam}.

Besides scalarization, there are also Pareto-based methods, such as evolutionary algorithms (EA) \cite{zhang2007moea, deb2002fast}, hypervolume-maximization approaches \cite{beume2007sms, zhang2023hypervolume}, and Pareto-based MOBO \cite{lin2022pareto}, which produce a set of Pareto-optimal points in each run rather than a single point as in scalarization methods. These approaches also avoid the challenging task of selecting an appropriate preference vector for each objective to reflect the true user preference. However, since preference vectors are absent in the formulation of Pareto-based methods, there is no direct control over the user preference, potentially leading to wasted computation on regions of little interest on the Pareto front.

Considering the benefits of STCH, which allows gradient-based optimization and provides direct control over the user preference, we adopt STCH for our multi-objective optimization framework. To approximate the full Pareto front, we aim to find the optimal model parameter $\boldsymbol{\theta}^*$ such that $\boldsymbol{x}^*(\boldsymbol{w})=h_{\boldsymbol{\theta}^*}(\boldsymbol{w})$ is the corresponding Pareto optimal solution for any given preference vector $\boldsymbol{w}$, where $h$ is the surrogate model that maps any valid preference vector $\boldsymbol{w}$ to its corresponding Pareto optimal solution. We describe in detail the surrogate model used in Section \ref{sec: 4.3}.

\subsection{Comparison with Existing Work} \label{sec: 2.3}

In addition to the research gaps mentioned at the end of each subsection above, we want to further articulate our work as a new methodological contribution rather than an incremental improvement. See table \ref{table:comparison with exsiting work} for a summary of the comparison with the existing work.

\begin{table*}[t]
\centering
\caption{Comparison with Existing Work}
\label{table:comparison with exsiting work}
\setlength{\tabcolsep}{6pt}
\renewcommand{\arraystretch}{1.15}
\begin{tabularx}{\textwidth}{l l l l l}
\toprule
Existing work & Architecture & Type of Sparsity & Limitations & SparseST Improvement (ours) \\
\midrule
\makecell[l]{Sparse Convolution \cite{graham2017submanifold}} &
CNN &
Spatial only &
\makecell[l]{-- Active sites dilation issue \\ -- No temporal sparsity} &
\makecell[l]{-- DN algorithm to mitigate dilation \\ -- Exploits temporal correlation} \\
\addlinespace[15pt]
\makecell[l]{DN Algorithm \cite{neil2017delta}} &
RNN/GRU &
Temporal only &
\makecell[l]{-- Fixed delta threshold $\Theta$ \\ -- No spatial sparsity} &
\makecell[l]{-- Learnable delta threshold $\Theta$ \\ -- Leverages sparse/fixed background} \\
\addlinespace[15pt]
\makecell[l]{Spartus \cite{gao2022spartus}} &
LSTM on FPGA &
\makecell[l]{Weight pruning \\ + temporal} &
\makecell[l]{-- Speech/audio only \\ -- Compresses model weights \\ -- No spatial data sparsity} &
\makecell[l]{-- Applied to video prediction/anomaly detection \\ -- No capacity loss from compression \\ -- Tunable performance--efficiency trade-off} \\
\bottomrule
\end{tabularx}
\end{table*}

\section{Preliminary Background} \label{sec: 3}

In this section, we review the basics of the ConvLSTM network and introduce the concepts of the spatially sparse tensor.

\subsection{Convolutional LSTM network}

A ConvLSTM unit is composed of an input gate $i_t$, a forget gate $f_t$ and an output gate $o_t$, indexed by time step $t$.
The spatial feature of the input tensor at the current time step $\mathcal{X}_t$ and hidden state $\mathcal{H}_{t-1}$ from the last time step are firstly extracted by the convolution filters with their corresponding weights. Each of the three gates $i_t$, $f_t$, $o_t$ at each time step, with a range of values between 0 and 1, acts as a valve to control how much information should be passed through. The update equations for a single LSTM unit are as follows \cite{shi2015convolutional}:

\begin{equation} \label{eq:convLSTM update}
   \begin{aligned}
i_t & =\sigma\left(\mathcal{W}_{x i} * \mathcal{X}_t+\mathcal{W}_{h i} * \mathcal{H}_{t-1}\right) \\
f_t & =\sigma\left(\mathcal{W}_{x f} * \mathcal{X}_t+\mathcal{W}_{h f} * \mathcal{H}_{t-1}\right) \\
o_t & =\sigma\left(\mathcal{W}_{x o} * \mathcal{X}_t+\mathcal{W}_{h o} * \mathcal{H}_{t-1}\right) \\
\tilde{\mathcal{C}}_t & =\tanh \left(\mathcal{W}_{x c} * \mathcal{X}_t+\mathcal{W}_{h c} * \mathcal{H}_{t-1}\right) \\
\mathcal{C}_t & =f_t \odot \mathcal{C}_{t-1}+i_t \odot \tilde{\mathcal{C}}_t \\
\mathcal{H}_t & =o_t \odot \tanh \left(\mathcal{C}_t\right)
    \end{aligned}
\end{equation}

where $\mathcal{W}$ denotes the weight matrices of convolutional filters, and $\sigma$, $\tanh$, $\odot $, $\ast$ represent the Sigmoid activation function,  Hyperbolic Tangent activation function, element-wise multiplication, and convolution operator, respectively.

\subsection{Spatially Sparse Tensor}

\begin{figure}[!tb]
    \centering
    \includegraphics[width=3.4in]{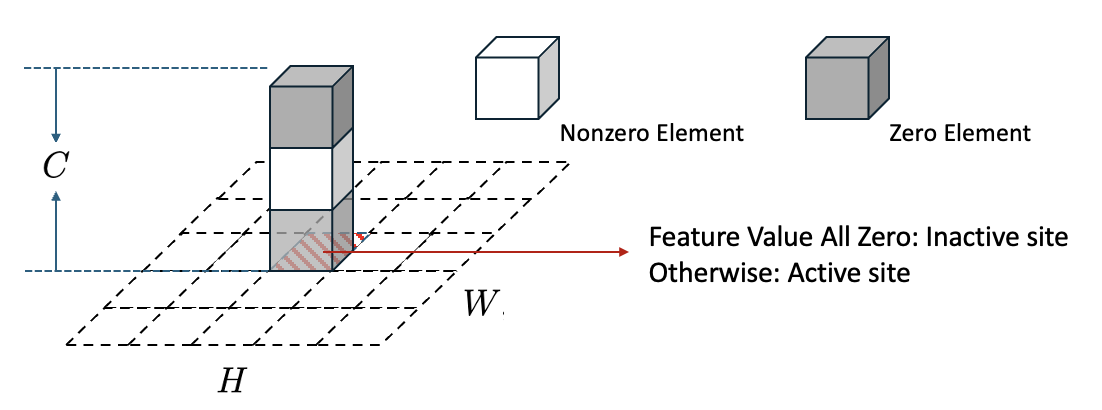}
    \caption{Example of an active site. Each spatial location on the grid is an inactive site if all the feature values of the column vector located at this site are zero. Otherwise, it is defined as an active site \cite{graham2017submanifold}.}
    \label{fig:sparse tensor}
\end{figure}

Following the notation in \cite{graham2017submanifold}, an input tensor has $(d+1)$ dimensions, where $d$ is the spatial dimension with an additional feature dimension. Figure \ref{fig:sparse tensor} illustrates an example of an active site on a grid with spatial dimension $H$ and $W$, in which case $d=2$. 
 
A spatially sparse tensor denotes a tensor with spatial locations that are mostly inactive sites, which can be represented by two matrices: a coordinate matrix to store the spatial locations of all active sites, and a feature matrix to store the corresponding feature values. For input tensor $\mathcal{T} \in \mathbb{R}^{C_{in} \times H\times W }$ with the spatial dimension $H\times W$ and feature dimension $C_{in}$, we can represent it as follows:

\begin{equation}
\mathcal{L}=\left[\begin{array}{cc}
h_1 & w_1 \\
\vdots&\vdots \\
h_N & w_N  
\end{array}\right], \mathcal{F}=\left[\begin{array}{c}
F_1^T \\
\vdots \\
F_N^T
\end{array}\right]
\label{eq:sparse tensor}
\end{equation}

where $N$ is the total number of active sites, and $F_i \in \mathbb{R}^{C_{in}}$ is the column feature vector for $i$-th active site.

\section{Methodology: SparseST Framework} \label{sec: 4}

In this section, we first introduce the preliminary knowledge on 2D sparse convolution and the delta network algorithm. Then, the unit structure and full model architecture of our proposed SparseST are introduced to demonstrate how 2D Sparse Convolution can be integrated with the delta algorithm to leverage both the spatial and temporal sparsities from the input data. In addition, we give the computational cost analysis of the proposed model. Finally, to explicitly control the trade-off between model performance and efficiency, we explain how to formulate the multi-objective optimization problem and design the composite loss function.

\subsection{2D Sparse Convolution}

In \cite{graham20183d}\cite{graham2017submanifold}, two types of Spatially Sparse Convolution are introduced and compared: Sparse Convolution and Submanifold Sparse Convolution. The latter preserves the same level of sparsity by preventing active sites in feature maps from expanding during forward propagation across network layers. However, this constraint limits our framework to extrapolate new active sites. Therefore, this type of convolution is inherently inapplicable to spatiotemporal data mining. To avoid this limitation, we adopt the 2D Sparse Convolution in our framework, which discards the convolution calculation of inactive sites and reduces the computational cost compared with the dense convolution operation \cite{graham20183d}\cite{graham2017submanifold}. 

For input tensor $\mathcal{T} \in \mathbb{R}^{C_{in}\times H\times W}$ in the form of Equation (\ref{eq:sparse tensor}), the output tensor $\mathcal{Y}$ at spatial location $(u, v)$ can be represented as:

\begin{equation}
   \mathcal{Y}_{u,v}=\begin{cases}
    \underset{(u+i, v+j)\in R}{\sum}\mathcal{W}_{\delta}^\top F_{u+i, v+j} & \text{if $(u+i, v+j) \in \mathcal{L}$}.\\
    \qquad \quad0 & \text{otherwise}
  \end{cases}
  \label{eq:sparseconv2d}
\end{equation}

% change (p, q) to (u+i, v+j), C to R under summation, 

where $F_{u+i, v+j}\in  \mathbb{R}^{C_{in}}$ is the column feature vector at input spatial location $(u+i, v+j)$, $\mathcal{W}_{\delta}^\top \in \mathbb{R}^{C_{out}\times C_{in}}$ is the transpose of corresponding filter weight partition and $R$ is a receptive field coordinate set.

\begin{figure}[!tb]
    \centering
    \includegraphics[width=3.4in]{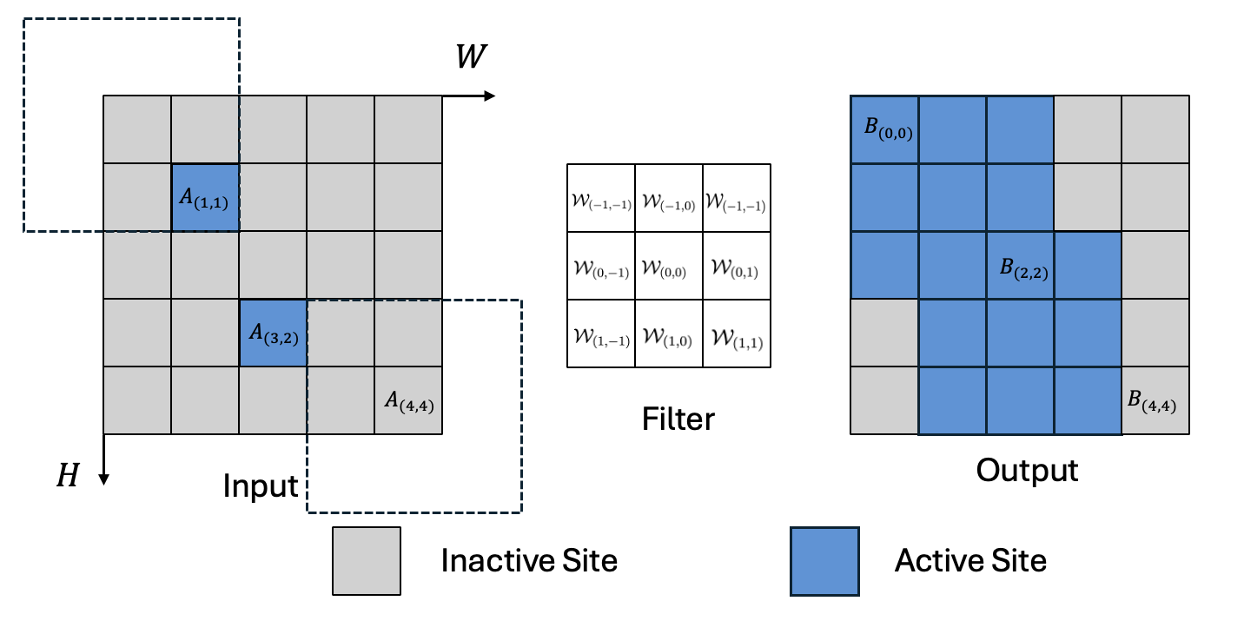}
    \caption{An example of 2D Sparse Convolution. Active and inactive sites in the input and output feature maps are represented as blue and gray blocks, respectively. Input feature map is zero-padded to keep the spatial size unchanged. When the receptive field of the convolutional filter (dotted frame) is at the top left corner, there is only one active site $A_{(1, 1)}$, yielding the corresponding output $B_{(0, 0)} = \mathcal{W}_{(1, 1)}\times A_{(1, 1)}$, discarding MAC computations of all other inactive sites. We can finally get the output feature map following the same principle as described above.}
    \label{fig: Sparseconv2d}
\end{figure}

Figure \ref{fig: Sparseconv2d} shows an example of 2D Sparse Convolution. It is noteworthy that the number of active sites colored in blue in the output is getting larger compared with that of the input, which means 2D Sparse Convolution will dilate the number of active sites, therefore reducing the computational efficiency in the following layers. To mitigate the dilation issue of 2D Sparse Convolution, as well as to leverage the data or feature sparsity introduced by temporal correlation, we introduce the general formulation of the DN Algorithm in the next part, followed by our proposed framework, SparseST.

\subsection{Delta Network Algorithm}

% add a logic summary of A, B, C 

% $x$ for scalar $\mathbf{x}$ for vector $X$ for matrix $\mathcal{X}$ for tensor

% Check the redundancy of character

Consider a multiplication  of a weight matrix $\mathbf{W}$ with an input vector sequence $\mathbf{X} = \{\mathbf{x_1}, \mathbf{x_2}, ..., \mathbf{x_T}\}$ where the component vector $\mathbf{x_t}$ is indexed by time step $\mathbf{t}$, the output vector sequence $\mathbf{Y} = \{\mathbf{y_1}, \mathbf{y_2}, ..., \mathbf{y_T}\}$ can be calculated in a recursive way:

\begin{equation} \label{eq:general DN algorithm}
    \begin{aligned}
         \mathbf{y_0} &= \mathbf{0}, \mathbf{x_0} = \mathbf{0} \\
         \Delta \mathbf{x_t} &= \mathbf{x_t} - \mathbf{x_{t-1}} \\
         \mathbf{y_t} &= \mathbf{W}\mathbf{x_t} = \mathbf{W} \Delta \mathbf{x_t} + \mathbf{y_{t-1}}
    \end{aligned}
\end{equation}

where the delta vector $\Delta \mathbf{x_t}$ is the difference of two input sequence elements from adjacent time steps and $\mathbf{y_{t-1}}$ is the multiplication result from the last time step, which is stored in the delta memory.

In the delta vector update Equations (\ref{eq:delta update}) \cite{neil2017delta}, we set a delta threshold $\Theta$ to zero out the below-threshold elements in the delta vector $\Delta \mathbf{x_t}$. Therefore, the multiplication of entire columns in a weight matrix with zero elements in the delta vector can be skipped. 

\begin{equation}
\begin{aligned}
\hat{x}_{i, t-1} & = \begin{cases}x_{i, t-1} & \text { if }\left|x_{i, t}-\hat{x}_{i, t-1}\right|>\Theta \\
\hat{x}_{i, t-2} & \text { otherwise }\end{cases} \\
\Delta x_{i, t} & = \begin{cases}x_{i, t}-\hat{x}_{i, t-1} & \text { if }\left|x_{i, t}-\hat{x}_{i, t-1}\right|>\Theta \\
0 & \text { otherwise }\end{cases}
\end{aligned}
\label{eq:delta update}
\end{equation}

where $\Delta x_{i, t}$ is the $i$-th element in the delta vector $\Delta \mathbf{x_t}$ at time step $t$, $\hat{x}_{i, t-1}$ is used to prevent the accumulation of delta approximation error. For example, when the absolute difference between $x_{i, t}$ and $\hat{x}_{i, t-1}$ is less than the threshold $\Theta$, the $\Delta x_{i, t}$ will be set to zero, and therefore introducing an approximation error less than $\Theta$. To prevent the accumulation of this approximation error, we need to update $\hat{x}_{i, t-1}$ to reset the approximation error when there is an update above the threshold.

\subsection{SparseST Unit Structure} \label{sec: 4.1}

\begin{figure}[!tb]
    \centering
    \includegraphics[width=3.4in]{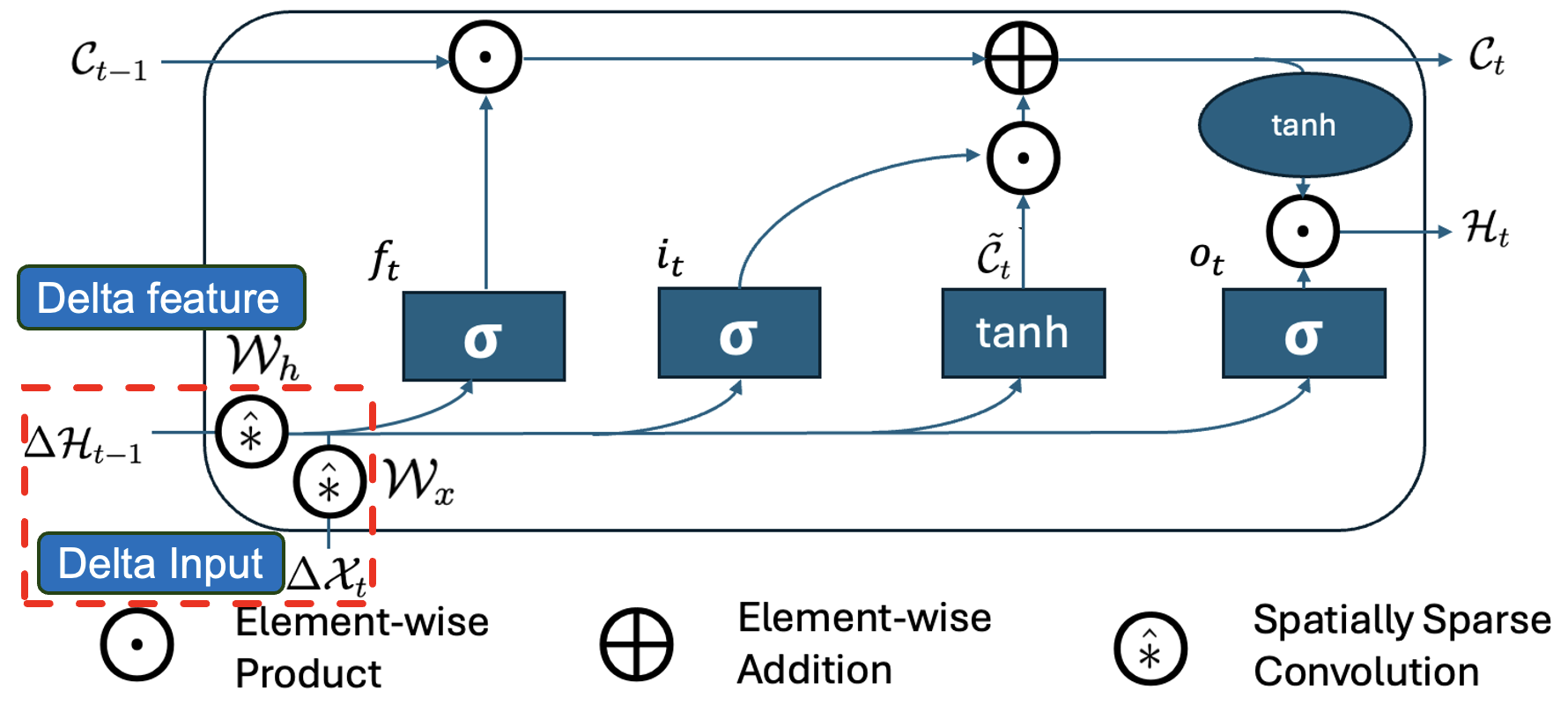}
    \caption{Unit structure for our proposed SparseST model.}
    \label{fig: SparseST model structure}
\end{figure}

The proposed unit structure of our SparseST framework is built by innovatively integrating sparse convolution and the delta algorithm into the ConvLSTM unit, which is shown in Figure \ref{fig: SparseST model structure}. It aims to fully exploit the spatial and temporal sparsity to reduce the computational complexity. In our proposed model, the delta tensors $\Delta \mathcal{X}_t$ and $\Delta \mathcal{H}_{t-1}$ for both the input and hidden states are generated as the difference between two adjacent time steps. And then the 2D Sparse Convolution is leveraged to skip computations on inactive sites following the example in Figure \ref{fig: Sparseconv2d} .

When updating the three gates $i_t$, $f_t$, $o_t$ and memory cell $\mathcal{C}_t$ at time step $t$, we can recursively get the result by keeping a record of previous sum-product results stored in the delta memory $M_{t-1}$ for each update. The update equation for a single SparseST unit is shown in Equation (\ref{eq:SparseST unit update}).

\begin{equation}
\begin{aligned}
\{i, f, o\}_t &= \sigma (M_{t}^{(g)}) \quad g \in \{i,f,o\} \\&= \sigma(\mathcal{W}_x^{(g)}   
 \;\hat{*}\;\Delta \mathcal{X}_t
        + \mathcal{W}_h^{(g)} \;\hat{*}\; \Delta \mathcal{H}_{t-1}
        + M_{t-1}^{(g)})\\
\tilde{\mathcal{C}}_t
&= \tanh(M_{t}^{(c)}) \\
&= \tanh\!\big(
   \mathcal{W}_{x}^{(c)} \;\hat{*}\; \Delta \mathcal{X}_t
 + \mathcal{W}_{h}^{(c)} \;\hat{*}\; \Delta \mathcal{H}_{t-1}
 + M_{t-1}^{(c)} \big)\\
\mathcal{C}_t &= f_t \odot \mathcal{C}_{t-1} + i_t \odot \tilde{\mathcal{C}}_t\\
\mathcal{H}_t &= o_t \odot \tanh(\mathcal{C}_t)
\end{aligned}
\label{eq:SparseST unit update}
\end{equation}

% where $\Delta \mathcal{X}_t = \mathcal{X}_t - \mathcal{X}_{t-1} $ and $\Delta \mathcal{H}_{t-1} = \mathcal{H}_{t-1} - \mathcal{H}_{t-2}$ are the delta tensors for the input and hidden state, respectively. $M_{G, t-1}$ are delta memory for input gate, forget gate, output gate and memory candidate, which store the sum-product result from the previous time step for computational reuse, respectively.

The delta tensor update equation for $\Delta \mathcal{X}_t$ and $\Delta \mathcal{H}_{t}$ as shown in Equation (\ref{eq:delta update for x and h}), where $\delta$ represents the delta thresholding function defined in Equation (\ref{eq:delta update}). Note that all of the delta thresholds for each unit are learnable and optimized along with model weights during training.

\begin{equation}
\begin{aligned}
\Delta \mathcal{X}_t &= \delta(\mathcal{X}_t - \mathcal{X}_{t-1}, \Theta_\mathcal{X}) \\
\Delta \mathcal{H}_t &= \delta(\mathcal{H}_t - \mathcal{H}_{t-1}, \Theta_\mathcal{H})\\
\end{aligned}
\label{eq:delta update for x and h}
\end{equation}

See Algorithm \ref{algo:SparseST unit update} for the detailed update flow in a single SparseST unit.

\begin{algorithm}[ht] 
\caption{SparseST Unit Update Flow} \label{algo:SparseST unit update}
\DontPrintSemicolon

\Input{Training data sequence $\mathcal{X} = \{\mathcal{X}_1, \mathcal{X}_2, ..., \mathcal{X}_T\}$}

\Output{Trained unit parameter $\boldsymbol{\theta} = \{\mathcal{W}_{x},\mathcal{W}_h, \Theta_\mathcal{X}, \Theta_\mathcal{H}\}$}

% \Initialize{Delta memories  $M_{i, 0}$, $M_{f, 0}$, $M_{o, 0}$, $M_{c, 0}$}

\BlankLine
\BlankLine

\For{t in iterations}{
    Calculate delta tensors $\Delta \mathcal{X}_t $ and $\Delta \mathcal{H}_{t-1}$ by Equation (\ref{eq:general DN algorithm})

    Process $\Delta \mathcal{X}_t$ and $\Delta \mathcal{H}_{t-1}$ using 2D Spatially Sparse Convolution by Equation (\ref{eq:sparseconv2d})
    
    Update the gates, cell state and delta memories in the unit by Equation 
    (\ref{eq:SparseST unit update})

    Update delta tensors using delta thresholding by Equation (\ref{eq:delta update for x and h})

    Update model parameter $\boldsymbol{\theta}$ by Backpropagation through Time (BPTT) \cite{werbos1990backpropagation}   
    }

\Return {Trained unit parameter $\boldsymbol{\theta}$}
\end{algorithm}

In our design, Sparse Convolution reduces computation by skipping inactive sites but dilates active spatial sites across layers. The DN algorithm can suppress redundant temporal information by creating delta tensors and mitigate the dilation issue by thresholding. Their hybridization ensures that both spatial and temporal redundancies are minimized: spatially through zero skipping and temporally through delta thresholding.

\subsection{Computational Cost Analysis} \label{sec: 4.2}

In this section, we analyze the computational cost of a SparseST unit and compare it with the baseline ConvLSTM unit. 

Table \ref{table:spconv2d flops} shows the computational cost comparison between 2D dense convolution and Sparse Convolution. The acceleration ratio (AR), which measures the reduction in computational cost, is defined as the spatial sparsity of the input tensor, which is the ratio of inactive sites.

\begin{table}[h]
    \centering
    \begin{tabular}{ll|ccc}
\hline \textbf{Dense Convolution} & \textbf{Sparse Convolution} & \textbf{Acceleration Ratio} \\
\hline \quad $HWK^2C_{in}C_{out}$ & \qquad $DK^2C_{in}C_{out}$ & $1-D/HW$ \\
\hline
\end{tabular}
    \caption{Computational cost comparison between 2D dense convolution and Sparse Convolution, where $H$ and $W$ are the spatial size of input tensor, $K$ is the size of convolutional filter, $D$ is the number of active sites in the input tensor, $C_{in}$ and $C_{out}$ are the number of channels of input and output tensors, respectively. }
    \label{table:spconv2d flops}
\end{table}

In Equation (\ref{eq:convLSTM update}) of a single ConvLSTM unit, the total computational cost for gate updates is as follows:

\begin{equation}
    FLOP_{gate} = 4\times HWK^{2}C_{out}(C_{in} + C_{out})
\end{equation}
 
The update of cell state $\mathcal{C}_t$ involves $2 * HWC_{out}$ number of FLOPs for element-wise multiplication and $HWC_{out}$ number of FLOPs for element-wise addition. The hidden state $\mathcal{H}_t$ involves $HWC_{out}$ number of FLOPs for element-wise multiplication. Therefore, the total computational cost for state updates is as follows: 

\begin{equation}
    FLOP_{states} = 4\times HWC_{out}
\end{equation}

The total computational cost for a single ConvLSTM cell is as follows:

\begin{equation}
    \begin{aligned}
        FLOP_{dense} & = FLOP_{gate} + FLOP_{states} \\
        & = 4\times HWC_{out}[K^2(C_{out}+C_{in})+1]
    \end{aligned}
\end{equation}

Note that we didn't include the computational cost for activation functions (sigmoid and tanh function) for simplicity.

For the computational cost of a single SparseST unit, we follow the same analysis as the ConvLSTM unit and combine the result in Table \ref{table:spconv2d flops}. Therefore, we have the following:

\begin{equation}
    \begin{aligned}
        FLOP_{sparse} & = 4\times DK^{2}C_{out}(C_{in} + C_{out}) + 4\times HWC_{out} \\
        & = 4*C_{out}[DK^2(C_{out}+C_{in})+HW]
    \end{aligned}
\end{equation}

The acceleration ratio of a single SparseST unit compared with the baseline ConvLSTM unit can be calculated as follows:

\begin{equation} \label{eq:AR calculation}
    \begin{aligned}
        AR & = (FLOP_{dense} - FLOPs_{sparse})/FLOP_{dense} \\
        & = (HW-D)/(HW+\epsilon) \\
        & \approx 1 - D/HW
    \end{aligned}
\end{equation}

where $\epsilon = 1/K^{2}(C_{in}+C_{out})$ is negligible compared with $HW$ in the above denominator. Note that the result in Equation (\ref{eq:AR calculation}) is exactly the ratio of inactive sites (spatial sparsity) of the inputs for a single unit. For a tiny numerical example, if the delta thresholds $\Theta_\mathcal{X}$ and $\Theta_\mathcal{H}$ introduce the spatial sparsity of 60\% and 40\% in $\Delta \mathcal{X}_t$ and $\Delta \mathcal{H}_{t-1}$, respectively, we report the AR defined in Equation (\ref{eq:AR calculation}) as 50\%, which is the average ratio of spatial sparsity for both delta tensors in one SparseST unit.

\subsection{Composite Loss for Multi-objective Optimization}\label{sec: 4.3}

From Section \ref{sec: 4.1}, we have two sets of parameters to optimize during training. One is the weight matrices $\mathcal{W}$ in Equation \ref{eq:SparseST unit update}. Another is the delta threshold $\Theta$ in Equation \ref{eq:delta update for x and h}, which directly controls the number of active sites $D$ in Equation \ref{eq:AR calculation} for computing AR. A larger $\Theta$ corresponds to a smaller $D$ for a higher AR, since we introduce more sparsity in the delta tensor. However, we achieve this higher efficiency at the cost of model performance.

Therefore, we consider model performance and efficiency to be two conflicting optimization objectives. We aim to enable practitioners to set their preferences under different settings of computational resources, thereby balancing the trade-off between these two objectives. 

First, the formulations of these two objectives are defined as follows: the mean square error (MSE) computed as the average squared difference between the predicted and ground truth pixel values across all channels and spatial dimensions for each data sample $\mathcal{L}_{\mathrm{MSE}}(\boldsymbol{\theta})$, and the average unit occupancy computed as the average of the ratio of active sites across all units in the model $\mathcal{L}_{\mathrm{Occupancy}}(\boldsymbol{\theta})$.

\begin{equation}
    \mathcal{L}_{\mathrm{MSE}}(\boldsymbol{\theta})=\frac{1}{n} \sum_{i=1}^n\left(y_i-f\left(x_i ; \boldsymbol{\theta}\right)\right)^2
\end{equation}

\begin{equation}
    \mathcal{L}_{\mathrm{Occupancy}}(\boldsymbol{\theta})=\frac{1}{NHW}\sum^{N}\sum^{H}\sum^{W}\mathbf{\mathds{1}(\mathcal{T}_{\boldsymbol{\theta}})}
\end{equation}

The indicator function for counting the active sites can be defined as follows:
\begin{equation}
\mathds{1}(\mathcal{T}_{\boldsymbol{\theta}}) =
\begin{cases}
0 & \text{if } \mathcal{T}_{:, i, j}(\boldsymbol{\theta}) = 0 \\
1 & \text{otherwise}
\end{cases}
\end{equation}
where $\boldsymbol{\theta}$ is the network parameter, including model weights and learnable delta thresholds,  $N$ is the total number of units, $H$ and $W$ are the spatial dimensions for tensor $\mathcal{T}\in \mathbb{R}^{C \times H \times W}$. In our case, the tensor $\mathcal{T}$ can be either delta input $\Delta \mathcal{X}_t$ or delta feature $\Delta \mathcal{H}_{t-1}$.

Given the two objectives defined above, we formulate the problem into a multi-objective optimization (MOO) problem and solve it with the Smooth Tchebycheff Scalarization method \cite{lin2024smooth} considering the benefits that (1) it allows gradient-based optimization method with modern deep learning optimizer; (2) it provides direct control over the user preference towards each objective; (3) it has a complete theoretical guarantee to explore the Pareto optimal points, including those on the non-convex part of Pareto front.

\begin{proposition}(Smooth Tchebycheff Scalarization Method)

For a general MOO problem: 
    \begin{equation} \label{eq:original MOO problem}
\min _{\mathbf{x}\in \mathbf{X}}  \mathbf{F}(\mathbf{x})=\left[F_1(\mathbf{x}), F_2(\mathbf{x}), \cdots, F_k(\mathbf{x})\right]^T
\end{equation}
where $\mathbf{x} \in \mathbb{R}^n$ is a vector of design variables, $\mathbf{X}$ is the feasible design space, and $\mathbf{F}(\mathbf{x}) \in \mathbb{R}^k$ is a vector of objective functions $F_i(\mathbf{x}): \mathbb{R}^n \rightarrow{} \mathbb{R}$.

It can be transformed as the following optimization problem by the smooth Tchebycheff scalarization method:
\begin{equation} \label{eq:STCH method}
g_\mu^{(\mathrm{STCH})}(\mathbf{x} \mid \boldsymbol{w})=\mu \log \left(\sum_{i=1}^k e^{\frac{w_i\left(f_i(\mathbf{x})-z_i^*\right)}{\mu}}\right)
\end{equation}

where $\boldsymbol{w} \in \mathbb{R}^k$ is the preference vector, and $w_i \geq 0, 
\sum_{i=1}^k w_i=1$ for all $i$, $z_i^*$ is the ideal value for each objective, $\mu>0$ is the smoothing factor.
\end{proposition}

During model training, we minimize the following composite loss function to minimize the MSE loss $\mathcal{L}_{\mathrm{MSE}}$ and minimize the average unit occupancy $\mathcal{L}_{\mathrm{Occupancy}}$, thereby maximizing the average unit sparsity for efficiency.
\begin{equation}
\label{eq:composite loss}
\begin{split}
g_{\mu}^{(\mathrm{STCH})}(\boldsymbol{\theta}\mid\boldsymbol{w})
&= \mu \log\!\Big(\exp\!\big(\tfrac{w_{\mathrm{MSE}}\big(\mathcal{L}_{\mathrm{MSE}}(\boldsymbol{\theta})-z_1^*\big)}{\mu}\big)\\[-2pt]
&+ \exp\!\big(\tfrac{(1-w_{\mathrm{MSE}})\big(\mathcal{L}_{\mathrm{Occupancy}}(\boldsymbol{\theta})-z_2^*\big)}{\mu}\big)
\Big).
\end{split}
\end{equation}

where $w_{MSE}$ is the preference weight on the MSE loss, which provides a direct control for users to balance the trade-off between model performance and efficiency.

\begin{proposition} (Ability to Find All Pareto Solutions, quoted from \cite{lin2024smooth}) \label{proposition: Proposition 2}
\\
 Under mild conditions, there exists a $\mu^*$ such that, for any $0<$ $\mu<\mu^*$, every Pareto solution of the original multi-objective optimization problem \eqref{eq:original MOO problem} is an optimal solution of the Smooth Tchebycheff scalarization problem \eqref{eq:STCH method} with some valid preference $\boldsymbol{\lambda}$.
\end{proposition}

See the mild conditions \cite{lin2024smooth} in Appendix B. These two propositions indicate that optimizing the loss function in Equation (\ref{eq:composite loss}) can recover Pareto optimal points in both convex and non-convex regions of the Pareto front. Consequently, the remaining challenge is to find a surrogate model $h$, described in Section \ref{sec: 2.2}, to explore and approximate the entire Pareto front efficiently.

Lin et al. \cite{lin2024smooth} employed a two-layer multi-layer perceptron (MLP) as the surrogate model. However, in our setting, acquiring objective values is computationally expensive since it requires fully training a deep neural network until convergence for each preference vector to obtain a single Pareto optimal point. This high cost makes it impractical to generate the large number of Pareto optimal points needed to train a neural network surrogate model effectively. To efficiently explore the Pareto front with minimal rounds of neural network training, we adopt a multi-task GP regression framework. Specifically, we train a GP as the surrogate model with a vector output to approximate the mapping from preference weights to Pareto-optimal outcomes. We begin by initializing the GP with a set of Pareto optimal points obtained by training the neural network within a predefined range of preference weights. At each iteration, an acquisition function based on the sum of marginal variance selects the next preference weight to explore. The neural network is then trained with this new preference, yielding a new Pareto optimal solution that is used to update the GP. At the end of the training loop, the trained GP can be queried with arbitrary preference weights to generate an approximated Pareto front, which is presented in Section \ref{sec: 5}.

\section{Experiments} \label{sec: 5}

In this section, we design two experiments on two different downstream tasks to present the effectiveness of our framework. One is to perform a spatiotemporal prediction task on the Moving MNIST dataset. The other is anomaly detection task on real industrial dataset.

\subsection{Case Study: Spatiotemporal Prediction}

\subsubsection{Dataset}
\begin{figure}[ht]
    \centering
    \includegraphics[width=3.4in]{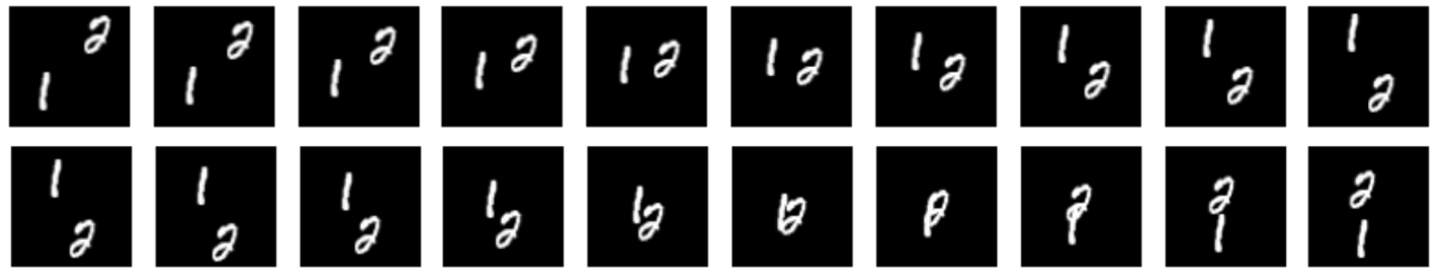}
    \caption{A sample of the Moving MNIST dataset.}
    \label{fig:datasample}
\end{figure}

\begin{figure*}[ht]
    \centering
    \subfloat[Moving digits ``7'' and ``8''.]{
        \includegraphics[width=0.8\textwidth]{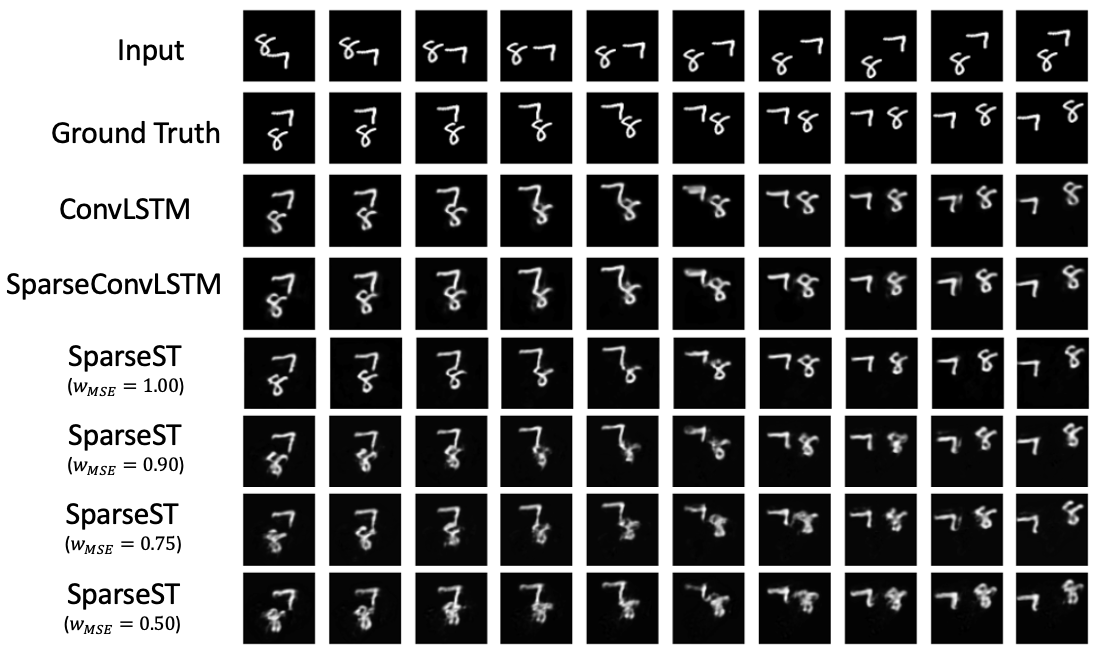}
        
    }
    \hfill
    \subfloat[Moving digits ``3'' and ``7''.]{
        \includegraphics[width=0.8\textwidth]{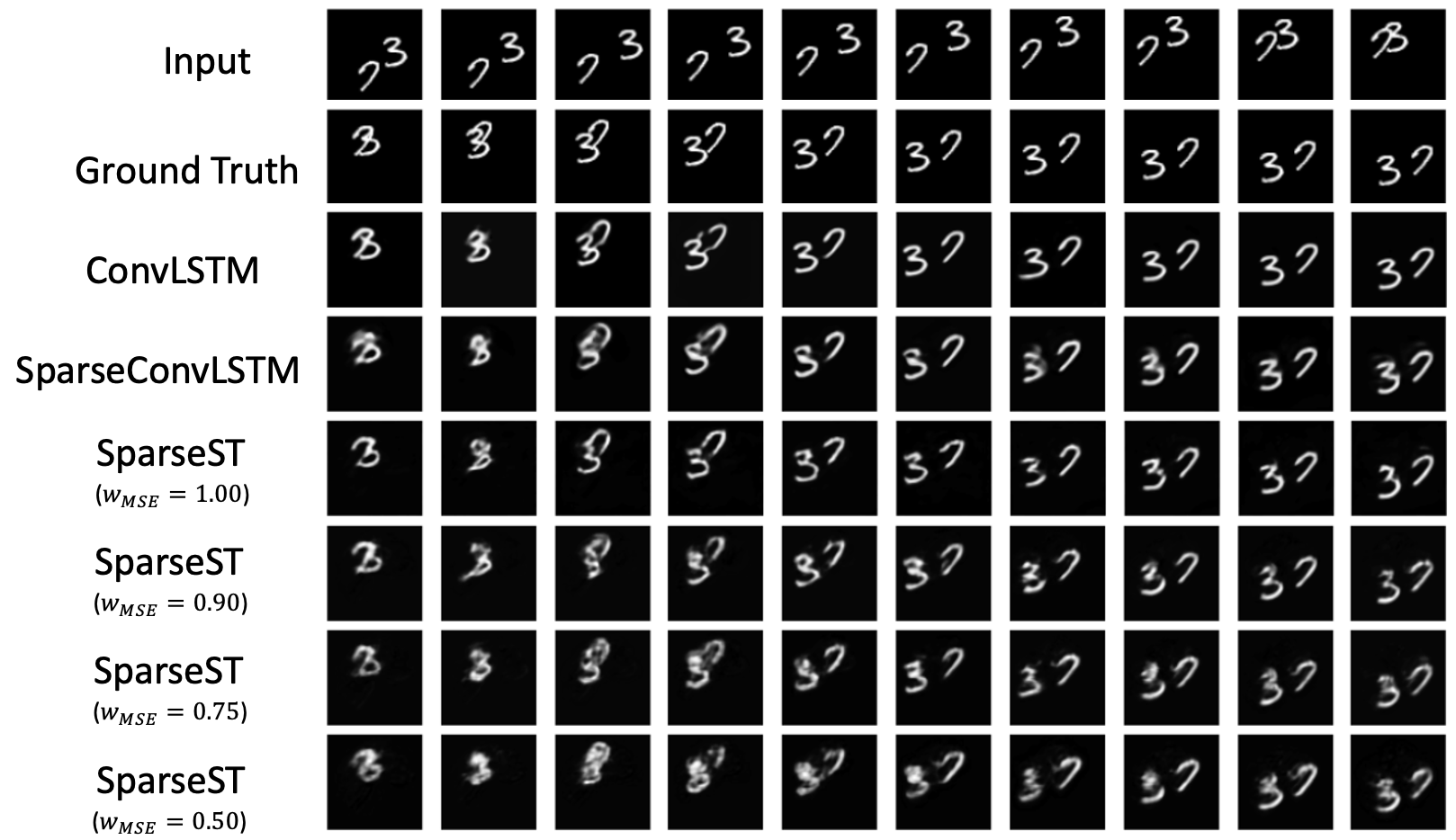}
    
    }
    \caption{Visualizations of model prediction results on Moving MNIST dataset.}
    \label{fig:mnist prediction visualization}
\end{figure*}

The Moving MNIST dataset \cite{srivastava2016unsupervised} is a synthetic spatiotemporal benchmark dataset, which contains sequences of moving handwritten digits. It is widely used for evaluating sequence modeling and video prediction models. Each data sample in this dataset has 20 grayscale images of size 64 $\times$ 64. Each image contains two digits of size 28 $\times$ 28, which is randomly chosen from the original MNIST dataset \cite{lecun1998gradient}, bouncing inside the 64 $\times$ 64 frame with a trajectory following a trigonometric pattern. The initial starting position of each digit is randomly picked inside the frame, and the velocity direction is randomly chosen on a unit circle with a magnitude over a fixed range \cite{srivastava2016unsupervised}. If the moving digit touches the frame boundary, the direction of its velocity will be inverted, but the magnitude will remain unchanged. Figure \ref{fig:datasample} shows a sample of the Moving MNIST dataset. The first row shows the moving trajectory of the two digits "1" and "2" for the first ten time steps, and the second row shows the trajectory for the next ten time steps.

\subsubsection{Experimental Settings}

Our task is next-frame spatiotemporal prediction with an autoencoder structure supervisedly. And our training, validation, and testing datasets have 8000, 2000, and 10000 samples, respectively. During model training, for each data sample with a length of 20 frames, the encoder takes in the 1st to 10th frames, and the decoder outputs the 2nd to 11th frames, which means that the model is trained to reconstruct the 2nd to 10th frames and make a prediction on the 11th frame. During model testing, we load the trained model with the lowest validation loss and predict the 11th to 20th images for each sample recursively. 

Two baseline models are included for this case study. One is the original ConvLSTM, 
and another is SparseConvLSTM (ConvLSTM only with 2D Sparse Convolution). SparseConvLSTM is considered one of the baseline models due to the high spatial sparsity of the moving MNIST dataset. We train the baseline models by minimizing the MSE loss, and the proposed SparseST model by minimizing the proposed composite loss in Equation \ref{eq:composite loss}. All models are trained by BPTT \cite{werbos1990backpropagation} for 200 epochs with early stopping applied using a patience of 10. Optimization is conducted using the Adam optimizer implemented in PyTorch \cite{NEURIPS2019_9015}, with an initial learning rate of 
$10^{-4}$ and a decay rate of 0.5. 

\subsubsection{Evaluation Metrics}

We compare our proposed model with the baseline methods by MSE loss for the model performance, and the analytical computational savings using the AR shown in Equation (\ref{eq:AR calculation}). We report the average AR over all SparseST units during testing.

\subsubsection{Results}

Figure \ref{fig:mnist prediction visualization} shows a visualization of the predicted results in two randomly selected data samples. For each sample, the first and second rows consist of the first ten frames as input for model training and the last ten frames as the ground truth, respectively. The following rows show the prediction results for different models. As we can see, ConvLSTM and SparseConvLSTM have comparable model performance, indicating that 2D Sparse Convolution achieves similar predictive accuracy to the standard dense Conv2D in the ConvLSTM structure. For SparseST, the model performance degrades when the preference weight $w_{MSE}$ decreases. This trend is expected, since setting a lower weight to the performance objective will shift the focus of optimization towards reducing the computational complexity, therefore achieving a higher acceleration rate at the expense of reduced predictive accuracy.

\begin{table*}[ht]
\centering
\setlength{\tabcolsep}{25pt}
\renewcommand{\arraystretch}{1.2}
\begin{tabular}{|c|c|c|c|}
\hline
Model   Type                                                          & MSE loss & GFLOPs & AR \\ \hline
\begin{tabular}[c]{@{}c@{}}ConvLSTM       \end{tabular} & $0.001715$  & 1322.88  & N/A       \\ \hline
SparseConvLSTM                                                        & $0.001607$  & 899.57   & 32.00\%   \\ \hline
SparseST    ($w_{MSE}=1.00$)                                   & $0.002750$   & 771.78  & 41.66\%  \\ \hline
SparseST    ($w_{MSE}=0.90$)                                    & $0.003859$   & 430.89  & 67.43\%    \\ \hline
SparseST    ($w_{MSE}=0.75$)                                     & $0.004067$  & 318.32  & 75.94\%    \\ \hline
SparseST    ($w_{MSE}=0.50$)                                     & $0.005739$  & 232.73   & 82.41\%    \\ \hline
SparseST    ($w_{MSE}=0.25$)                                     & $0.009493$  & 133.38   & 89.92\%    \\ \hline

SparseST    ($w_{MSE}=0.10$)                                     & $0.017967$  & 119.62   & 90.26\%    \\ \hline
\end{tabular}
\caption{Comparison of model performance, computational cost (GFLOPs) and acceleration ratio in different settings. GFLOPs is reported as the average number of FLOPs over the testing dataset during inference. For the parameter count, all models have the parameter size close to 7.71 MB. Our proposed SparseST model only introduces delta thresholds for each unit as additional parameters, which is negligible.}
\label{table:mnist_table}
\end{table*}

The numerical result in Table \ref{table:mnist_table} validates the above analysis. It is worth noting that SparseConvLSTM slightly outperforms ConvLSTM while achieving a 32\% reduction in computational cost. This suggests that on the Moving MNIST dataset, 2D Sparse Convolution effectively removes redundant computation caused by spatial sparsity. It preserves performance and even yields a slight improvement. In our proposed SparseST framework, we further exploit the temporal sparsity, which arises from subtle temporal changes of the informative pixels (moving digits) between adjacent time steps. By integrating the Delta Network algorithm, SparseST can leverage the temporal sparsity by optimizing the delta thresholds for each unit. As a result, our framework can achieve higher acceleration ratios, with the trade-off between model performance and acceleration controlled by the preference weight.

\subsection{Case Study: Anomaly Detection}

\subsubsection{Dataset}

The Industrial Process Anomaly Detection (IPAD) dataset \cite{liu2024ipad} is specifically curated for the video anomaly detection task in industrial manufacturing scenarios. It features a comprehensive collection of video sequences containing both real-world data acquired by live video recordings and simulated data generated by SolidWorks, a professional 3D modeling software. The dataset consists of 16 distinct types of industrial machinery, including but not limited to conveyor belts, lift tables, cutting machines, drilling machines, machine grippers, and cranes, providing a rich diversity of normal and abnormal operational patterns.

\subsubsection{Experimental Settings}

\begin{figure*}[!tb]
    \centering
    \includegraphics[width=7in]{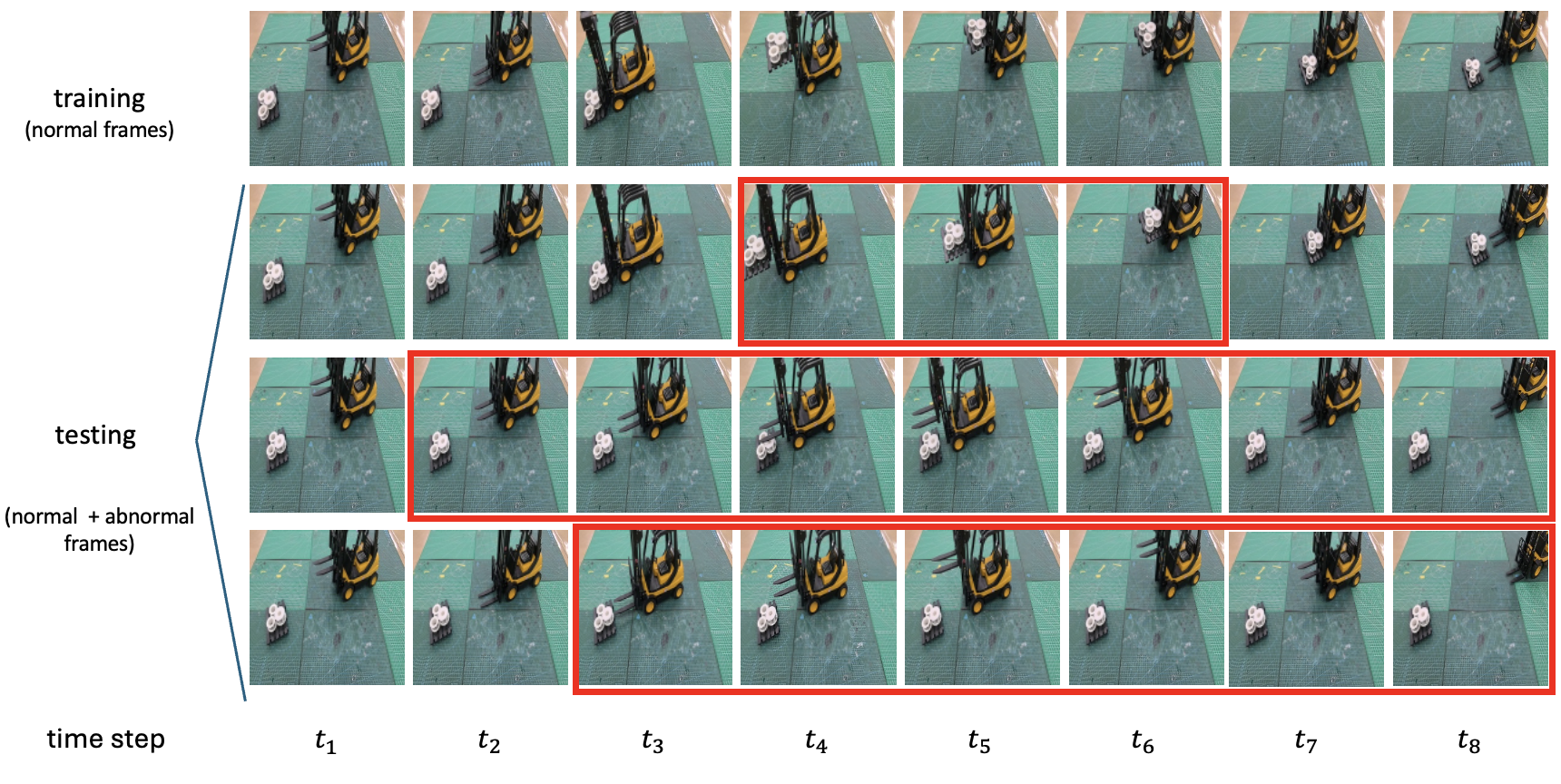}
    \caption{Sampled frames of a normal sequence. The first row shows a sample of frames of one training sequence; The last three rows show samples of frames of testing sequences with different types of anomaly. Frames highlighted in red are labeled as abnormal.}
    \label{fig: IPAD_sample}
\end{figure*}

In this case study, we select the crane scenario from the IPAD dataset, where a crane delivers a cargo from the bottom-left corner into the center of the frame. Each training sequence contains only normal frames, while testing sequences include both normal and abnormal segments with annotated frame-level labels. Each sequence consists of approximately 650 to 700 frames. To enhance the visibility of temporal dynamics between adjacent frames and considering the computational resource constraints, we sampled each sequence every 5 frames and downsampled the size of each frame to 3 $\times$ 64 $\times$64. Figure \ref{fig: IPAD_sample} shows several samples of frames in both training and testing sequences.

\begin{figure*}[!tb]
    \centering
    \subfloat[Training.]{
        \includegraphics[width=0.8\textwidth]{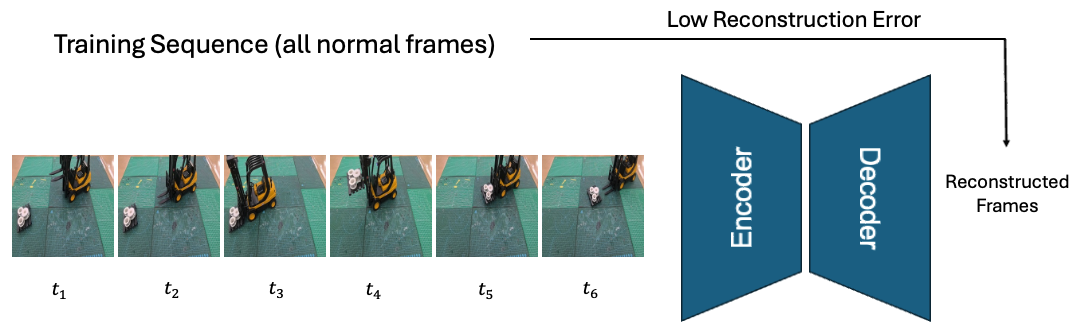}
        
    }
    \hfill
    \subfloat[Testing.]{
        \includegraphics[width=0.8\textwidth]{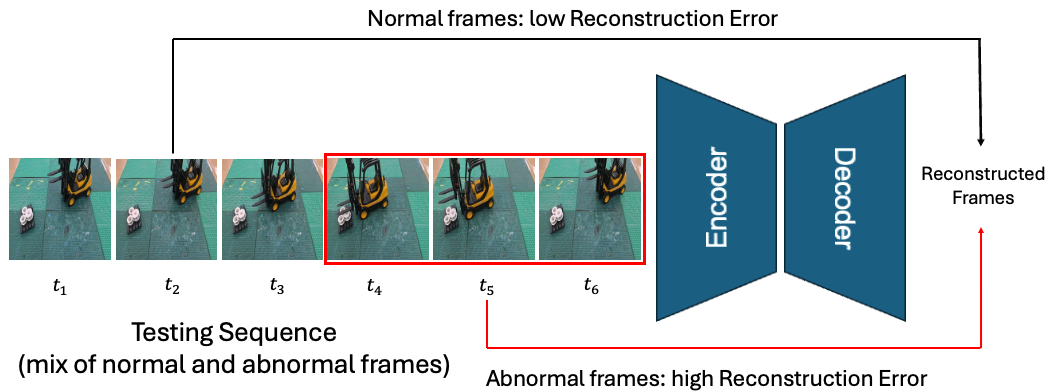}
    
    }
    \caption{Workflow of reconstruction-based anomaly detection.}
    \label{fig: anomaly_detection_flow_chart}
\end{figure*}

Figure \ref{fig: anomaly_detection_flow_chart} shows the flow chart of reconstruction-based anomaly detection. During training and validation, we adopt a sliding window approach with a window size of 21 frames and a stride of 1 to capture the long-range spatiotemporal dependencies within the sequence data. The model is trained to reconstruct all the input frames within each window. During testing, we only reconstruct the central frame of each sliding window and record the MSE between the reconstructed and ground truth frames as the anomaly score. As only normal frames are included in the training phase, in the testing phase, abnormal frames with features deviated from the normal patterns tend to yield higher reconstruction errors and thus, enable effective frame-level anomaly detection. All hyperparameters for training are kept the same as those in the previous experiments.

\subsubsection{Results}

\begin{figure*}[!tb]
    \centering
    \includegraphics[width=6in]{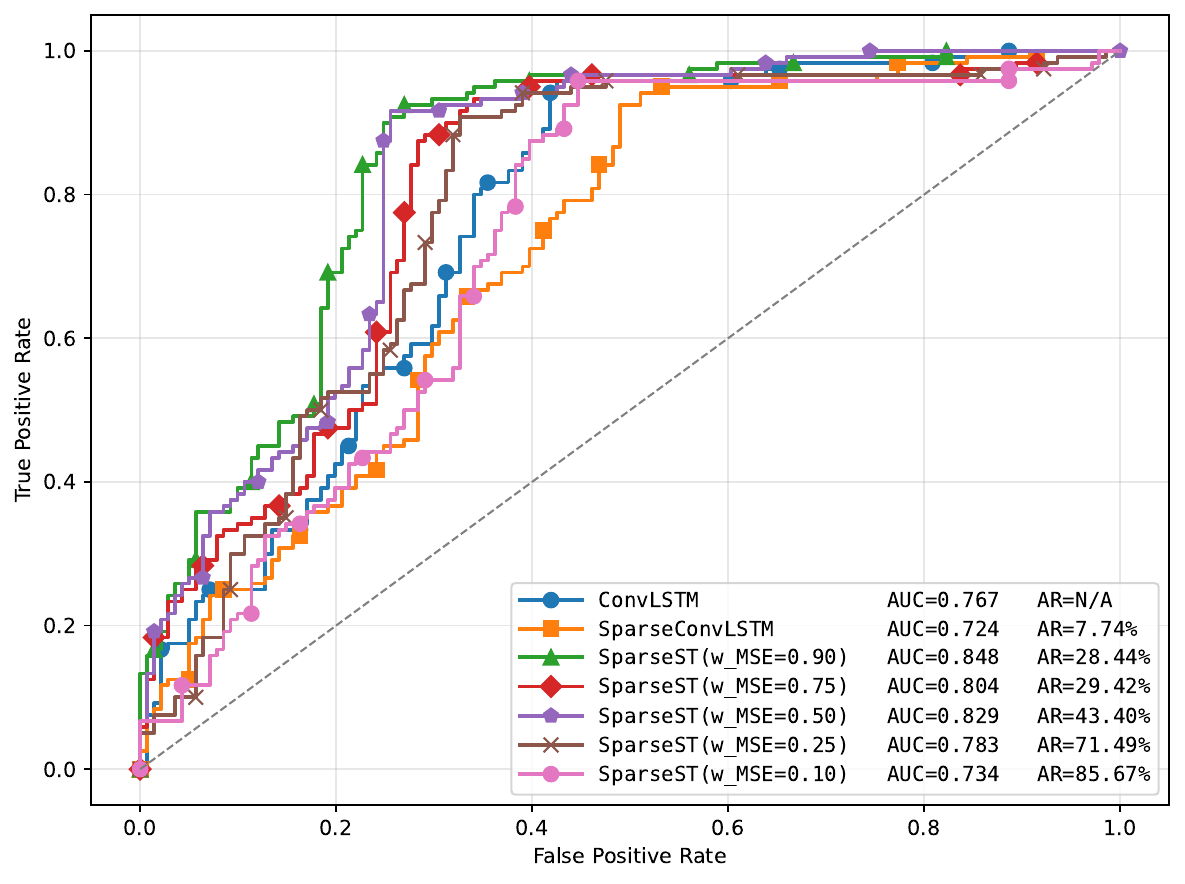}
    \caption{ROC curve comparison between different models.}
    \label{fig:IPAD_roc_curves}
\end{figure*}

Figure \ref{fig:IPAD_roc_curves} compares the receiver operating characteristic (ROC) curves of all models, along with their corresponding preference weight, AUC (area under the ROC curve), and acceleration ratio (AR) in the legend. 

We can observe that SparseConvLSTM model has a similar AUC score to the ConvLSTM model, and the AR is only 7.74\%. This is because the IPAD dataset consists of RGB images with little to no spatial sparsity compared with the Moving MNIST dataset. In this case, 2D Sparse Convolution itself cannot effectively leverage sparsity to improve efficiency without the DN algorithm. 

For the SparseST models, we integrate the DN algorithm to leverage the temporal sparsity of data, and the AR can be largely boosted. A lower preference weight of MSE introduces lower values of AUC (worse anomaly detection performance) and higher values of AR (lower computational costs), indicating greater efficiency at the expense of model performance. In particular, the SparseST models perform better (higher AUC scores) than the baseline model when the preference weight of MSE is greater than 0.25. This means, for reconstruction-based anomaly detection tasks, there is excess information within the data that can be traded for efficiency. Despite the spatiotemporal sparsity introduced by the framework, the information remained in the delta input or features is sufficient to enable the model to distinguish the normal pattern from the abnormal. However, when the preference weight of MSE is decreased to 0.10, the performance of SparseST turns worse than the baseline ConvLSTM, implying that the loss of information introduced by spatial and temporal sparsity in the data stream is too significant to provide sufficient information for anomaly detection.

% anomaly score difference between normal and abnormal frames
\begin{figure*}[!tb]
    \centering
    \includegraphics[width=6.5in]{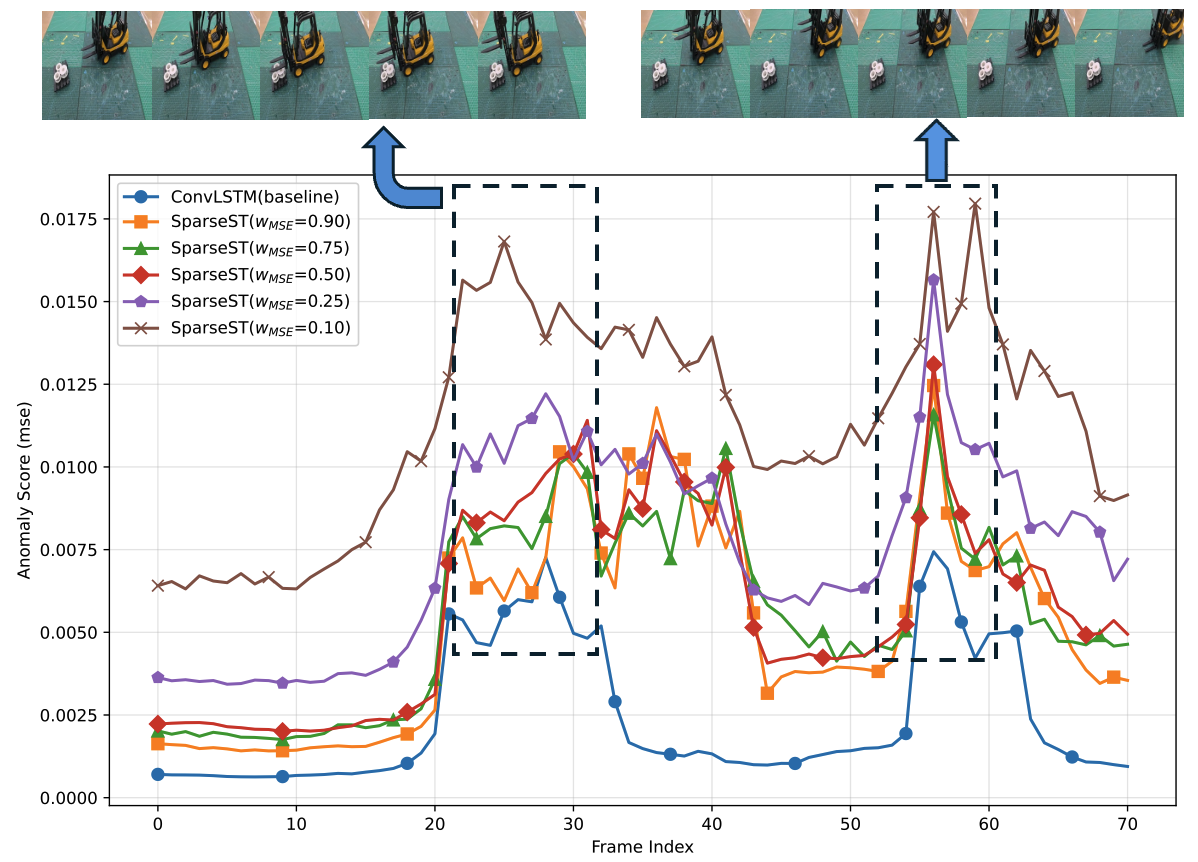}
    \caption{Difference of anomaly score between normal and abnormal frames.}
    \label{fig:IPAD_anomaly_score}
\end{figure*}

Figure \ref{fig:IPAD_anomaly_score} shows the anomaly score comparison between normal and abnormal frames using different models on a single testing sequence. The MSE for each frame in the sequence is plotted as the anomaly score. As highlighted in the figure, two distinct peaks appear consistently across all models, indicating that each model successfully detects two types of anomalies in this sequence: (1) when the crane approaches the cargo but fails to load it due to the arm being positioned too high, and (2) when the crane moves backwards to the destination, it fails to move the cargo with it. Although there is an increasing trend of anomaly scores across SparseST models when the preference weight for MSE decreases, the difference between anomaly scores on normal and abnormal frames is still significant enough for anomaly detection. This suggests that in this case study, our proposed framework can effectively trade redundant information in the data for efficiency.

\subsection{Pareto Front Analysis}

\begin{figure}[!ht]
\begin{centering}
\subfloat[Moving MNIST.]{\includegraphics[width=0.5\textwidth]{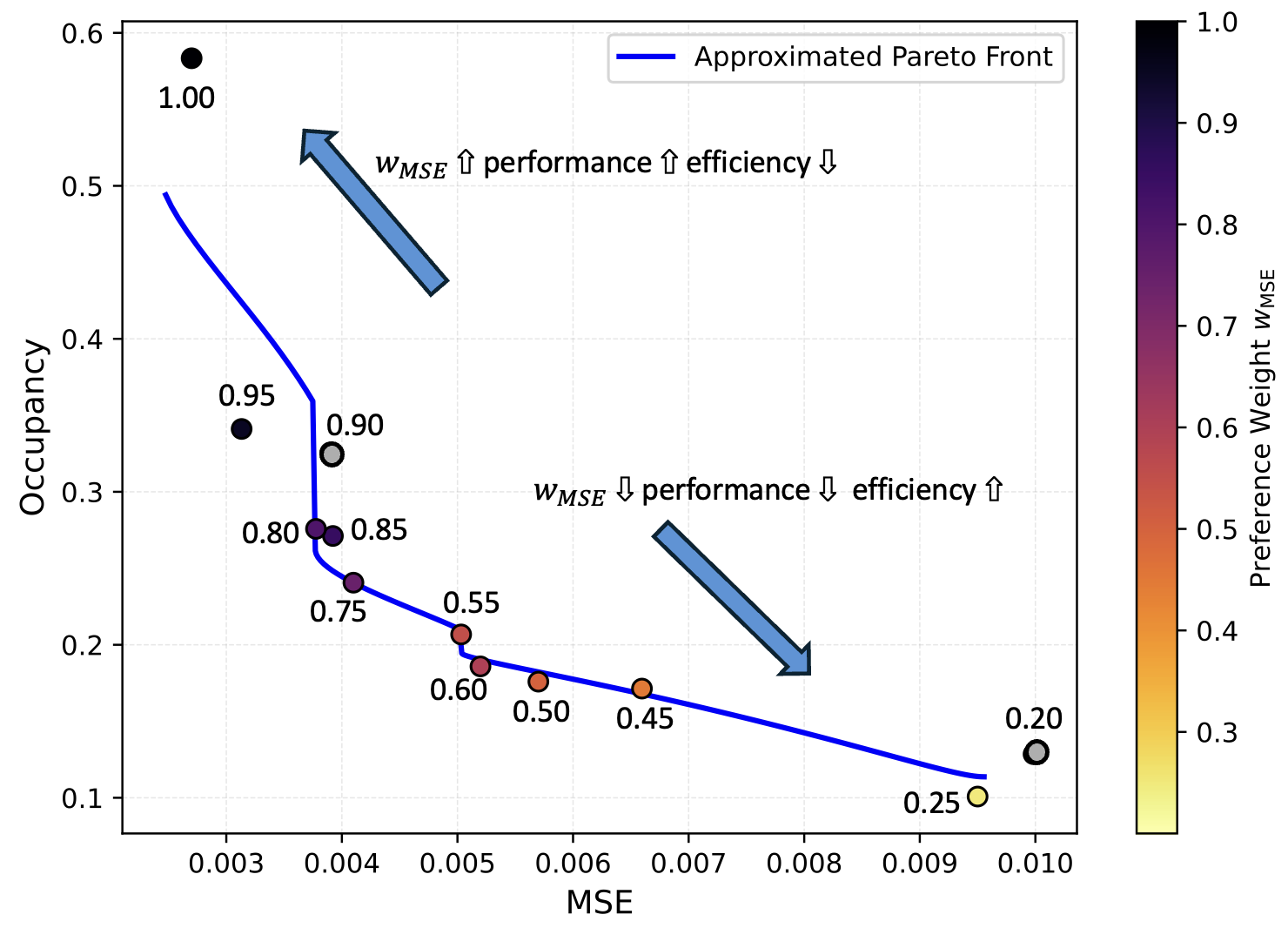}

}\par\end{centering}

\begin{centering}

\subfloat[IPAD.]{\includegraphics[width=0.5\textwidth]{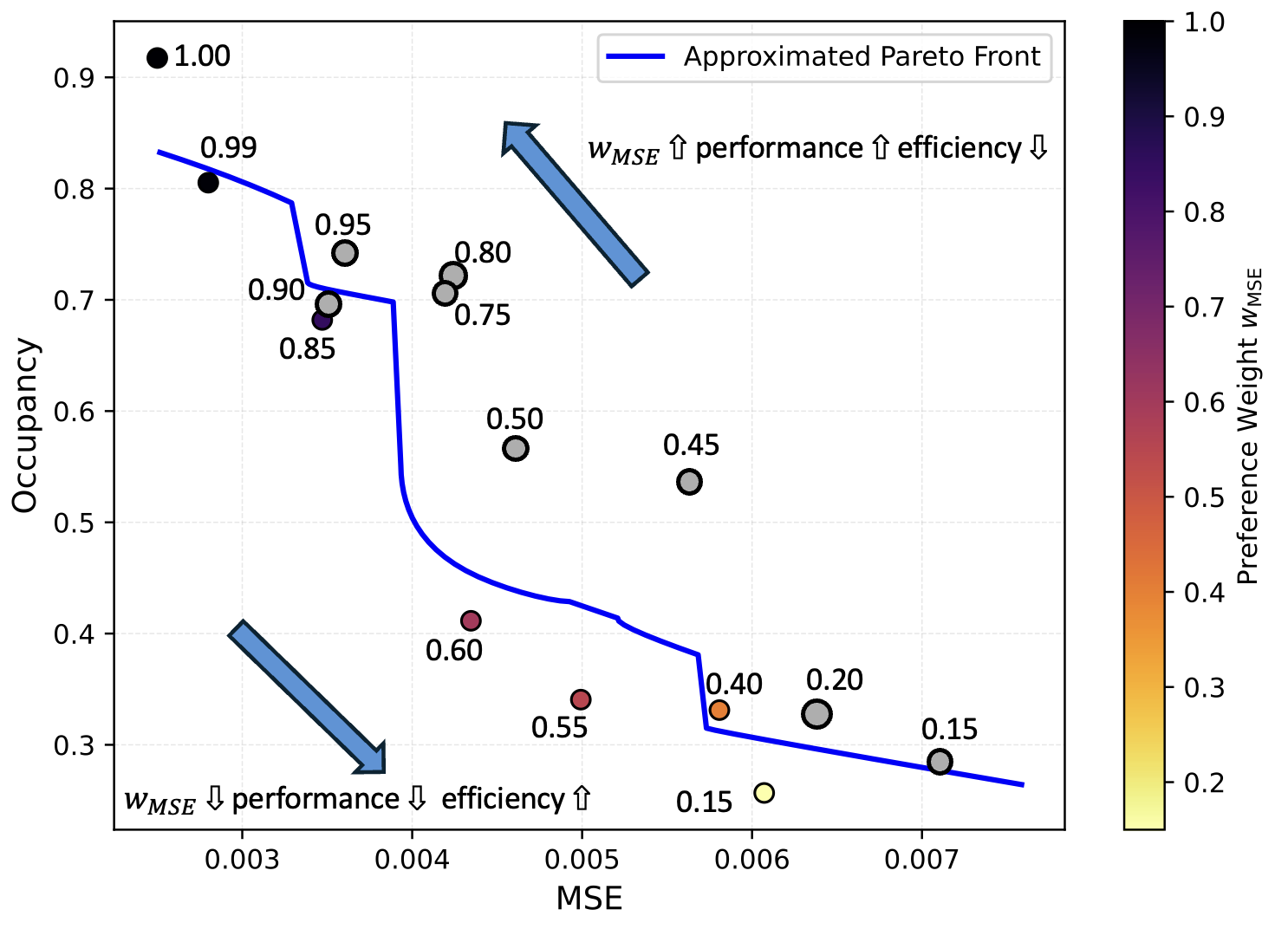}

}
\par\end{centering}

\caption{Visualization of Approximated Pareto Front. The points colored in gray are dominated, while non-dominated points are colored w.r.t the preference weight colorbar.}
\label{fig:pareto_front}
\end{figure}

Following the methodology mentioned in Section \ref{sec: 4.3}, we obtain 
Figure \ref{fig:pareto_front} to illustrate the approximated Pareto fronts for the two competing objectives: model performance on the downstream tasks (measured by MSE) and computational efficiency (measured by occupancy) in the above two case studies. They include both the initial points and the new candidate points with the largest sum of marginal variances of the two objectives in each iteration. Each point is annotated by the value of its associated preference weight $w_{MSE}$. Ideally, all of these points are supposed to be Pareto optimal (non-dominated). However, we obtain these points by training a neural network, which means the convergence and global optimality of the training process cannot be fully guaranteed. It is worth noting that the approximated Pareto fronts are non-convex, which empirically demonstrates the effectiveness of the smoothed Tchebycheff scalarization to recover Pareto-optimal solutions in non-convex regions of the objective space. And this is what standard linear scalarization fails to guarantee. A clear trade-off trend between the two objectives shows up along with the Pareto fronts. As the preference weight $w_{MSE}$ increases, the model achieves lower MSE (better accuracy) at the cost of higher occupancy (lower acceleration). Conversely, decreasing $w_{MSE}$ will favor sparsity and computational efficiency, but reduce predictive accuracy. This visualization can provide practical guidance for practitioners to select appropriate preference weights tailored to the computational resource constraints or accuracy requirements of a downstream task. For example, one can choose a point on the Pareto front that satisfies a specific range of MSE or occupancy, then finetune the preference weight according to the explored points. 

\section{Summary} \label{sec: 6}

In this work, we investigate efficient AI for STDM with the goal of exploiting spatial and temporal sparsity in raw data and intermediate features. Our motivation is to trade redundant information for efficiency, particularly in spatiotemporal data with fixed or slowly changing backgrounds, where modeling the full complex dependency is unnecessary.

To address this, we integrate 2D Sparse Convolution and the DN algorithm within a ConvLSTM backbone and propose the SparseST framework. This design not only improves computational efficiency but also overcomes key limitations: the DN algorithm mitigates the active site dilation problem of 2D Sparse Convolution and effectively leverages temporal correlations to exploit sparsity between consecutive frames.

Experimental results demonstrate that SparseST achieves substantial computational savings while maintaining accuracy comparable to baseline models. Moreover, for downstream tasks that do not rely on full spatiotemporal dependency modeling (e.g., anomaly detection), our framework provides improved efficiency without sacrificing or even improving the model performance. 

Furthermore, we formulate the performance–efficiency trade-off as a multi-objective optimization problem and design a composite loss function to balance the two objectives. By applying STCH scalarization within a multi-task GP regression, we approximate the Pareto front and provide practical guidance for adjusting preferences between computational cost and predictive performance, depending on downstream task requirements.

\section*{Acknowledgments}
This work was supported by the RPI-IBM Future of Computing Research Collaboration (FCRC) grant. We also thank all anonymous reviewers for their constructive comments.

\section*{Conflict of Interest}
All authors declare that they have no known conflicts of interest in terms of competing financial interests or personal relationships that could have an influence or are relevant to the work reported in this paper

\appendix
\section{Appendix A}\label{Appendix A}

\section{Pareto Optimality}
\begin{definition}\cite{ruchte2021scalable}
    Pareto dominance: A point $\mathbf{x}^* \in \mathbf{X}$ dominates another point $\mathbf{x'}$ when both: (1) $\mathbf{x}^*$ is not worse than $\mathbf{x'}$ on any objective, i.e. $F_i(\mathbf{x^*}) \leq F_i\left(\mathbf{x'}\right)$, $\forall i\in \{1, ..., k\}$; (2) $\mathbf{x}^*$ is better than $\mathbf{x'}$ on at least one objective, i.e. $\exists \;j\in \{1, ..., k\}$ s.t. $F_j(\mathbf{x^*}) < F_j\left(\mathbf{x'}\right)$.
\end{definition}

\begin{definition} \cite{ruchte2021scalable}
    Pareto optimality: A point $\mathbf{x}^* \in \mathbf{X}$ is Pareto optimal if it is not dominated by any other points. The Pareto frontier is the k-dimensional manifold of the objective values of all Pareto optimal solutions.
\end{definition}

\begin{definition} \cite{marler2004survey}
    Weak Pareto optimality: A point $\mathbf{x}^* \in \mathbf{X}$ is weakly Pareto optimal iff there does not exist another point $\mathbf{x} \in \mathbf{X}$, such that $F_i(x)<F_i(x^*)$, $\forall i\in \{1, ..., k\}$.
\end{definition}

\renewcommand{\theequation}{B\arabic{equation}}
\setcounter{equation}{1} 
\section{Assumptions for Proposition 2}

Define the Pareto front of a multi-objective optimization problem as a surface $\phi(x) = 0$, where $x\in\mathbb{R}^m$ is a vector in the objective space.

\begin{assumption}\cite{li1996convexification} For each point $x$ on the Pareto front $\phi$, assume:
\begin{itemize}
    \item $\nabla \phi(\boldsymbol{y}) \succ 0$, which means $\frac{\partial \phi(\boldsymbol{y})}{\partial y_i}>0$ for all $i=1, \ldots, m;$
    \item all second-order derivatives of $\phi(\boldsymbol{y})$ are bounded (i.e., the Hessian $\nabla^2 \phi(y)$ has finite eigenvalues).
\end{itemize}
\end{assumption}

The first statement assumes $\phi$ is strictly increasing in each objective, indicating that any weakly Pareto optimal solution must also be Pareto optimal. The second statement is a standard regularity assumption to ensure the scalarization is smooth enough.

\appendix
\section{Appendix B}\label{Appendix B}

\begin{figure}[!tb]
    \centering
    \includegraphics[width=3.4in]{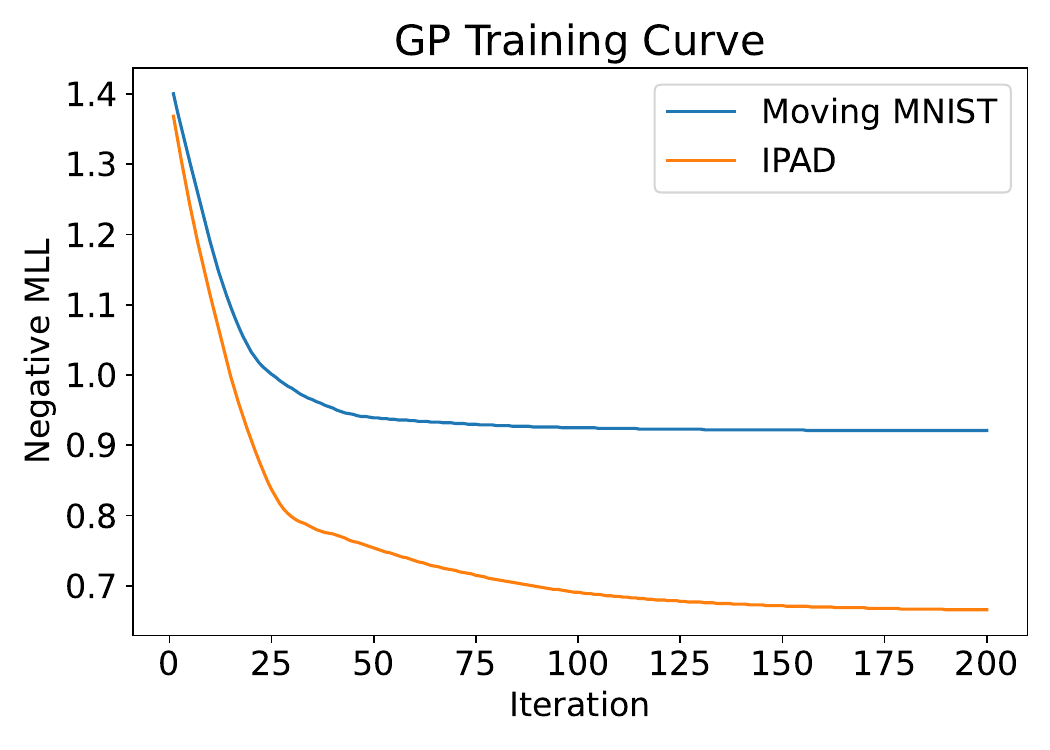}
    \caption{Change of negative marginal log likelihood (MLL) loss over GP training iterations.}
    \label{fig:GP_training_curve}
\end{figure}

The GP training curves for both experiments are shown in Figure \ref{fig:GP_training_curve}. In Figure, we show the predictions of the trained multi-task GP on each of the two objectives.  

\begin{figure}[!tb]
    \centering
    
    \subfloat[Moving MNIST.]{
        \includegraphics[width=3.5in]{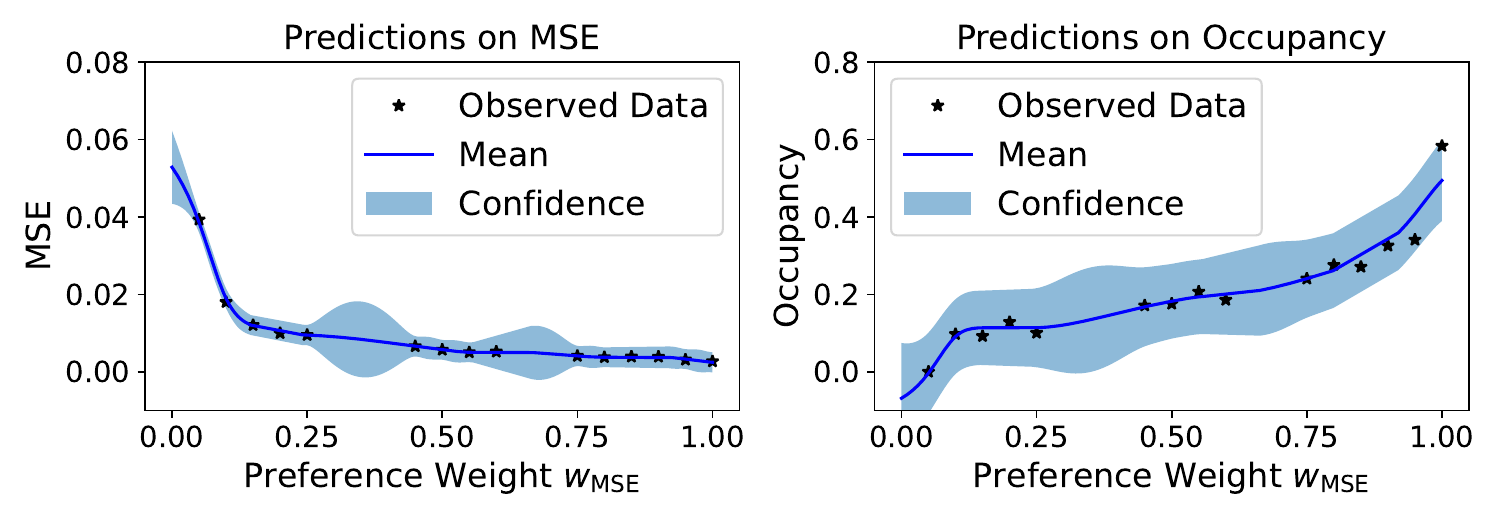}
        
    }
    \hfill
    \subfloat[IPAD.]{
        \includegraphics[width=3.5in]{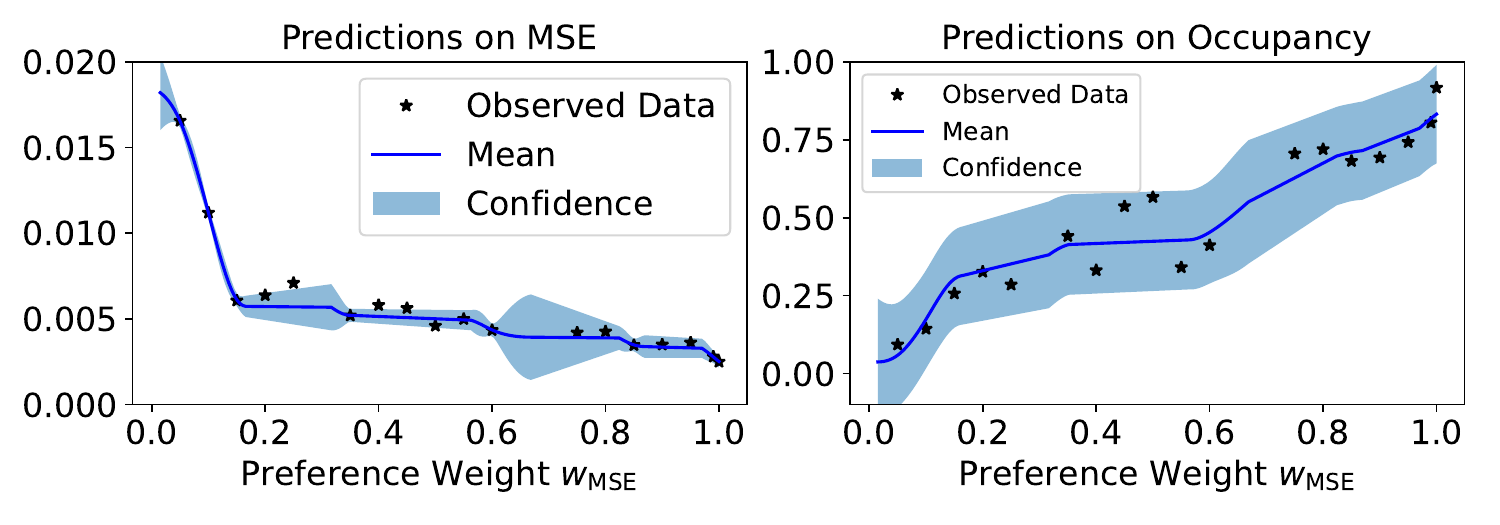}
    
    }
     \caption{Multi-task GP predictions on each objective for both experiments. The blue region is the 95\% confidence interval for each preference weight on the x-axis.}
    \label{fig:GP_predictions}
    
\end{figure}
 
\bibliographystyle{ieeetr}
\bibliography{bibliography}

\end{document}